\documentclass[preprint,12pt]{elsarticle}

\usepackage{graphicx}
\usepackage{subfigure}
\usepackage{amsmath}
\usepackage{amssymb}

\usepackage{verbatim}

\def\bi{\begin{itemize}}
\def\ei{\end{itemize}}
\def\bequ{\begin{equation}}
\def\eequ{\end{equation}}

\def\benum{\begin{enumerate}}
\def\eenum{\end{enumerate}}

\begin{document}

\begin{frontmatter}
%
% paper title
% can use linebreaks \\ within to get better formatting as desired
% Do not put math or special symbols in the title.
\title{Beyond Pixels: A Comprehensive Survey from Bottom-up to Semantic Image Segmentation and Cosegmentation}
%
%\author{authors}
%\markboth{xxx}%{xxx}
%\maketitle
%
\author[i2r]{Hongyuan Zhu}
\ead{zhuh@i2r.a-star.edu.sg}

\author[uestc]{Fanman Meng}
\ead{fmmeng@uestc.edu.cn}

\author[ntu]{Jianfei Cai\corref{cor1}}
\ead{asjfcai@ntu.edu.sg} \cortext[cor1]{Corresponding author.}

\author[i2r]{Shijian Lu}
\ead{slu@i2r.a-star.edu.sg}

\address[i2r]{Institute for Infocomm Research, A*STAR, Singapore}
\address[uestc]{School of Electronic Engineering, Univ. of Electronic Science and Technology, China}
\address[ntu]{School of Computer Engineering, Nanyang Technological University, Singapore}

\begin{abstract}
Image segmentation refers to the process to divide an image into
~non-overlapping meaningful regions according to human perception,
which has become a classic topic since the early ages of computer
vision. A lot of research has been conducted and has resulted in
many applications. However, while many segmentation algorithms exist, yet there are
only a few sparse and outdated summarizations available, an
overview of the recent achievements and issues is lacking. We aim
to provide a comprehensive review of the recent progress in this
field. Covering 180 publications, we give an overview of broad areas of segmentation topics including not only the classic bottom-up approaches, but also the recent development in superpixel, interactive methods, object proposals, semantic image parsing and image cosegmentation. In addition, we also review the existing influential
datasets and evaluation metrics. Finally, we suggest some design flavors and research
directions for future research in image segmentation.
\end{abstract}

% Note that keywords are not normally used for peerreview papers.
%\begin{IEEEkeywords}
\begin{keyword}
Image segmentation, superpixel, interactive image segmentation,
object proposal, semantic image parsing, image cosegmentation.
\end{keyword}
%\end{IEEEkeywords}

\end{frontmatter}

\section{Introduction}
Human can quickly localize many patterns and automatically group
them into meaningful parts. Perceptual grouping refers to human's
visual ability to abstract high level information from low level
image primitives without any specific knowledge of the image
content. Discovering the working mechanisms under this ability has
long been studied by the cognitive scientists since 1920's. Early
gestalt psychologists observed that human visual system tends to
perceive the configurational whole with rules governing the
psychological grouping. The hierarchical grouping from low-level
features to high level structures has been proposed by gestalt
psychologists which embodies the concept of grouping by proximity,
similarity, continuation, closure and symmetry. The highly compact
representation of images produced by perceptual grouping can
greatly facilitate the subsequent indexing, retrieving and processing.

With the development of modern computer, computer scientists
ambitiously want to equip the computer with the perceptual
grouping ability given many promising applications, which lays
down the foundation for image segmentation, and has been a
classical topic since early years of computer vision. Image
segmentation, as a basic operation in computer vision, refers to
the process to divide a natural image into $K$ non-overlapped
meaningful entities (e.g., objects or parts). The segmentation
operation has been proved quite useful in many image processing
and computer vision tasks. For example, image segmentation has
been applied in image annotation~\cite{CarsonBGM02} by decomposing
an image into several blobs corresponding to objects. Superpixel
segmentation, which transforms millions of pixels into hundreds or
thousands of homogeneous
regions~\cite{AchantaSSLFS12,ComaniciuM02}, has been applied to
reduce the model complexity and improve speed and accuracy of some
complex vision tasks, such as estimating dense correspondence
field~\cite{LuYMD13}, scene parsing~\cite{KohliKT09} and body
model estimation~\cite{Mori05}.
\cite{GuLAM09,CarreiraS12,EndresH14} have used segmented regions
to facilitate object recognition, which provides better
localization than sliding windows. The techniques developed in
image segmentation, such as Mean Shift~\cite{ComaniciuM02} and
Normalized Cut~\cite{ShiM00} have also been widely used in other
areas such as data clustering and density estimation.

\begin{figure*}
    \begin{center}
    \includegraphics[width = 1\linewidth]{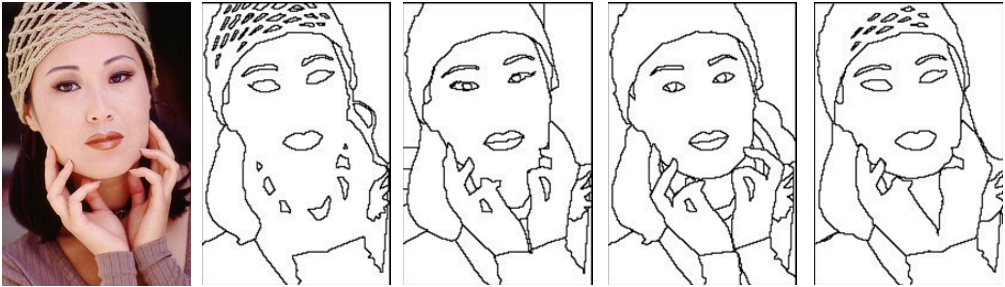}
    \end{center}
    \caption
    {\label{fig:illpose} An image from Berkeley segmentation dataset~\cite{MartinFM04} with hand labels from different human subjects, which demonstrates the variety of human perception.
    }
\end{figure*}

One would expect a segmentation algorithm to decompose an image
into the ``objects'' or semantic/meaningful parts. However, what makes
an ``object'' or a ``meaningful'' part can be ambiguous. An
``object'' can be referred to a ``thing'' (a cup, a cow, etc), a
kind of texture (wood, rock) or even a ``stuff'' (a building or a
forest). Sometimes, an ``object'' can also be part of other
``objects''. Lacking a clear definition of ``object'' makes
bottom-up segmentation a challenging and ill-posed problem.
Fig.~\ref{fig:illpose} gives an example, where different human
subjects have different ways in interpreting objects. In this
sense, what makes a `good' segmentation needs to be properly
defined.

%%What's segmentation
%----------------------------------------------------------------------
%------------------------

Research in human perception has provided some useful guidelines
for developing segmentation algorithms. For example, cognition
study~\cite{Hoffman1997} shows that human vision views part
boundaries at those with negative minima of curvature and the part
salience depends on three factors: the relative size, the boundary
strength and the degree of protrusion. Gestalt theory and other
psychological studies have also developed various principles
reflecting human perception, which include: (1) human tends to
group elements which have similarities in color, shape or other
properties; (2) human favors linking contours whenever the
elements of the pattern establish an implied direction.

Another challenge which makes ``object'' segmentation difficult is
how to effectively represent the ``object''. When human perceives
an image, elements in the brain will be perceived as a whole, but
most images in computers are currently represented based on
low-level features such as color, texture, curvature, convexity, etc. Such low-level features reflect local properties,
which are difficult to capture global object information. They are also sensitive to lighting and perspective
variations, which could cause existing algorithms to
over-segment the image into trivial regions.
%An example of the limitation of such low level cues is illustrated in Fig.~\ref{fig:InitialImpress}.
Fully supervised methods can learn higher level and global cues,
but they can only handle limited number object classes and require
per-class labeling.

A lot of research has been conducted on image segmentation. Unsupervised segmentation, as one classic topic in
computer vision, has been studied since 70's. Early
techniques focus on local region merging and
splitting~\cite{Ohlander78, Brice1970}, which borrow ideas from
clustering area. Recent techniques, on the other hand, seek
to optimize some global criteria~\cite{ComaniciuM02,
FelzenszwalbH04, ShiM00, VincentS91, Beare06, Chan2001, Osher88}.
Interactive segmentation methods~\cite{RotherKB04, YangCZL10,AnhCZZ12}
which utilize user input, have been applied in some commercial products
such as Microsoft Office and Adobe Photoshop. The substantial
development of image classification~\cite{ShottonWRC09}, object
detection~\cite{FelzenszwalbGM10}, superpixel
segmentation~\cite{AchantaSSLFS12} and 3D scene
recovery~\cite{HoiemEH08} in the past few years have boosted the
research in supervised scene parsing~\cite{KohliLT09, GouldFK09,
LiuYT11a, TigheL13, KrahenbuhlK11, LadickyRKT10}. With the
emergence of large-scale image databases, such as
the ImageNet~\cite{DengDSLL009} and personal photo streams on Flickr,
the cosegmentation methods~\cite{KimX12, KimXLK11, JoulinBP12,
RubioSLP12, JoulinBP10, ChangLL11, MukherjeeSP11, MukherjeeSD09}
which can extract recurring objects from a set of images, has
attracted increasing attentions in these years.

Although image segmentation as a community has been evolving for a long time, the
challenges ranging from feature representation to model design and
optimization are still not fully resolved, which hinder further
performance improvement towards human perception. Thus, it is necessary to
periodically give a thorough and systematic review of the segmentation
algorithms, especially the recent ones, to summarize what have been achieved, where we are
now, what knowledge and lessons can be shared and transferred
between different communities and what are the directions and
opportunities for future research. To our surprise, there are only some
sparse reviews on segmentation literature,
there is no comprehensive review which covers broad areas of
segmentation topics including not only the classic bottom-up approaches, but also the recent development in superpixel, interactive methods, object proposals, semantic image parsing and image cosegmentation, which will be critically and exhaustively reviewed in this paper in Sections~\ref{sec:bottom-up}-\ref{sec:coseg}. In addition, we will also review the existing influential
datasets and evaluation metrics in Section~\ref{sec:dataset}. Finally, we discuss some popular design flavors and some potential future directions in Section~\ref{sec:discuss}, and conclude the paper in Section~\ref{sec:summary}.

\section{Bottom-up methods}
\label{sec:bottom-up} The bottom-up methods usually do not take
into account the explicit notion of the object, and their goal is
mainly to group nearby pixels according to some local homogeneity
in the feature space, e.g. color, texture or curvature, by clustering
those features based on fitting mixture models, mode
shifting~\cite{ComaniciuM02} or graph
partitioning~\cite{FelzenszwalbH04}~\cite{ShiM00}\cite{VincentS91}\cite{Beare06}.
In addition, the variational~\cite{Chan2001} and level
set~\cite{Osher88} techniques have also been used in segmenting
images into regions. Below we give a brief summary of some popular
bottom-up methods due to their reasonable performance and publicly
available implementation. We divide the bottom-up methods into two
major categories: discrete methods and continues methods, where
the former considers an image as a fixed discrete grid while the
latter treats an image as a continuous surface.

\subsection{Discrete bottom-up methods}
\textbf{K-Means}:  K-means is among the simplest and
    most efficient method. Given $k$ initial centers which can be
    randomly selected, K-means first assign each sample to one
    of the centers based on their feature space distance and
    then the centers are updated. These two steps iterate
    until the termination condition is met. Assuming that cluster number is
    known or the distributions of the clusters are spherically
    symmetric, K-means works efficiently. However, these assumptions
    often don't meet in general cases, and thus K-means could have problems
    when dealing with complex clusters.

\textbf{Mixture of Gaussians}: This method and K-means are similar
to each other. Within mixture of Gaussians, each cluster center is now replaced by a covariance
    matrix. Assume a set of $d$-dimensional
    feature vector $x_1, x_2,..., x_n$ which are drawn from a Gaussian
    mixture:
    \begin{equation}
    p(x|\{\pi_k,\mu_k,\Sigma_k\})=\sum_{k}\pi_{k}N(x|\mu_k,\Sigma_k)
    \end{equation}
    where $\pi_k$ are the mixing weights, $\mu_k,\Sigma_k$ are the
    means and covariances, and
    \begin{equation}
    N(x|\mu_k,\Sigma_k)=\frac{exp\{-\frac{1}{2}(x-\mu_k)^T \Sigma_k^{-1}(x-\mu_k)\}}{(2\pi)^{d/2}|\Sigma_k|^{1/2}}
    \end{equation}
    is the normal equation~\cite{Szeliski11}. The parameters $\pi_k,\mu_k,\Sigma_k$
    can be estimated by using expectation maximization (EM) algorithm as
    follows~\cite{Szeliski11}.
    \begin{enumerate}
        \item The \textit{E-step} estimates how likely a sample $x_i$
        is
        generated from the $k$th Gaussian clusters with current
        parameters:
        \begin{equation}
        z_{ik}=\frac{1}{Z}N(x_i|\mu_k,\Sigma_k)
        \end{equation}
        with $\sum_{k}z_{ik}=1$ and $Z = \sum_{k} N(x_i|\mu_k,
        \Sigma_k)$.

        \item The \textit{M-step} updates the parameters:
        \begin{equation}
        \begin{split}
        &\mu_k=\frac{1}{n_k}\sum_{i}z_{ik}x_{i}, \\
        &\Sigma_k = \frac{1}{n_k}z_{ik}(x_{i}-\mu_k)(x_{i}-\mu_k)^T,\\
        &\pi_k = \frac{n_k}{n}
        \end{split}
        \end{equation}
        where $n_k = \sum_{i}z_{ik}$ estimates the number of sample points in each cluster.
    \end{enumerate}

    After the parameters are estimated, the segmentation can be formed
    by assigning the pixels to the most probable cluster. Recently, 
    Rao et al~\cite{RaoMYSM09} extended the gaussian mixture
    models by encoding the texture and boundary using minimum description length
    theory and achieved better performance. 

   \textbf{Mean Shift}: Different from the parametric methods such as K-means
    and Mixture of Gaussian which have assumptions over the cluster number
    and feature distributions, Mean-shift~\cite{ComaniciuM02}
    is a non-parametric method which can automatically decide the cluster number
    and modes in the feature space.

    Assume the data points are drawn from some probability function, whose density can be estimated by convolving the data with a fixed kernel of width
    $h$:
    \begin{equation}
    f(x)=\sum_{i}K(x-x_i)=\sum_{i}k(\frac{||x-x_i||^2}{h^2})
    \end{equation}
    where $x_i$ is an input sample and $k(.)$ is the kernel function~\cite{Szeliski11}.
    After the density function is estimated, mean shift uses a multiple restart gradient descent method which starts at
    some initial guess $y_k$, then the gradient direction of $f(x)$
    is estimated at $y_k$ and uphill step is taken in the direction~\cite{Szeliski11}. Particularly, the gradient of $f(x)$ is
    given by
    \begin{equation}
    \begin{split}
    \nabla f(x)& =\sum_i(x_i-x)G(x-x_i)\\
    &=\sum_i(x_i-x)g(\frac{||x-x_i||^2}{h^2})
    \end{split}
    \end{equation}
    where
    \begin{equation}
    g(r)=-k'(r)
    \end{equation}
    and $k'(.)$ is the first-order derivative of $k(.)$. $\nabla f(x)$ can be re-written
    as
    \begin{equation}
    \nabla f(x) = \left[ \sum_i G(x-x_i)\right] m(x)
    \end{equation}
    where the vector
    \begin{equation}
    m(x)=\frac{\sum_{i}x_iG(x-x_i)}{\sum_{i}G(x-x_i)}-x
    \end{equation}
    is called the mean shift vector which is the difference between the mean of the neighbors around $x$ and the current
    value of $x$. During the mean-shift procedure, the current mode $y_k$ is
replaced by its locally
    weighted mean:
    \begin{equation}
    y_{k+1}=y_k + m(y_k) .
    \end{equation}
    Final segmentation is formed by grouping pixels whose converge points are
    closer than $h_s$ in the spatial domain and $h_r$ in the range domain, and
    these two parameters are tuned according to the requirement of different
    applications.

\textbf{Watershed}: This method~\cite{VincentS91} segments
    an image into the catchment and basin by flooding the
    morphological surface at the local minimum
    and then constructing ridges at the places where different
    components meet. As watershed associates each region
    with a local minimum, it can lead to serious
    over-segmentation. To mitigate this problem,
    some methods \cite{Beare06} allow user to provide some initial
    seed positions which helps improve the result.

 \textbf{Graph Based Region Merging}: Unlike
    edge based region merging methods which use \textit{fixed} merging rules, Felzenszwalb
    and Huttenlocher \cite{FelzenszwalbH04} advocates a method which can use a \textit{relative}
    dissimilar measure to produce segmentation which optimizes a global grouping metric.
    The method maps an image to a graph with
    a $4$-neighbor or $8$-neighbor structure.
    The pixels denote the nodes, while the edge weights reflect the color
    dissimilarity between nodes. Initially each node forms their own component.
    The internal difference $Int(C)$ is defined as the largest weight in the
    minimum spanning tree of a component $C$. Then the weight is sorted in
    ascending order. Two regions $C_1$ and $C_2$ are merged if the in-between
    edge weight is less than
    $min(Int(C_1)+\tau(C_1),Int(C_2)+\tau(C_2))$, where $\tau(C)=k/|C|$ and $k$ is a coefficient that is used to control the component size.
    Merging stops when the difference between components exceeds the internal difference.
%   The author can prove that the resulting segmentation is neither too coarse
%   (there exist regions that could have been split) or too fine (there exist regions
%   that could have been merged).

\textbf{Normalized Cut}\label{subsec:ncut}: Many methods generate
the segmentation based on local image statistics only,
    and thus they could produce trivial regions because the low level features
    are sensitive to the lighting and perspective changes. In contrast, Normalized Cut~\cite{ShiM97}
    finds a segmentation via splitting the affinity graph which encodes the
    global image information, i.e. minimizing the \emph{Ncut} value between different clusters:
    \begin{eqnarray}
    Ncut(S_1,S_2,...,S_k):=\frac{1}{2} \sum_{i=1}^k
    \frac{W(S_i,\bar{S_i})}{vol(S_i)} \label{eq:NCUT}
    \end{eqnarray}
    where $S_1,S_2,...,S_k$ form a $k$-partition of a graph,
    $\bar{S_i}$ is the complement of $S_i$, $W(S_i,\bar{S_i})$ is the
    sum of boundary edge weights of $S_i$, and $vol(S_i)$ is the sum
    of weights of all edges attached to vertices in $S_i$. The basic
    idea here is that big clusters have large $vol(S_i)$ and
    minimizing \emph{Ncut} encourages all $vol(S_i)$ to be about the
    same, thus achieving a ``balanced" clustering.

    Finding the normalized cut is an NP-hard problem. Usually, an
    approximate solution is sought by computing the eigenvectors $v$ of
    the generalized eigenvalue system
    $(D-W)v=\lambda Dv$, where $W = [w_{ij}]$ is the affinity matrix of
    an image graph with $w_{ij}$ describing the pairwise affinity of
    two pixels and $D=[d_{ij}]$ is the diagonal matrix with
    $d_{ii}=\sum_{j}w_{ij}$.
    In the seminal work of Shi and Malik~\cite{ShiM97},
    the pair-wise affinity is chosen as the Gaussian kernel
    of the spatial and feature difference for pixels within a radius $||x_i-x_j||<r$:
    \begin{equation}\label{pair-wise affinity}
    W_{ij} = exp(-\frac{||F_i-F_j||^2}{\sigma_F^2}-\frac{||x_i-x_j||^2}{\sigma_s^2})
    \end{equation}
    where $F$ is a feature vector consisting of intensity, color and Gabor features
    and $\sigma_F$ and $\sigma_s$ are the variance of the feature and spatial position,
    respectively. In the later work of Malik \emph{et al.}~\cite{MalikBSL99}, they define a
    new affinity matrix using an intervening contour method. They measure the difference
    between two pixel $i$ and $j$ by inspecting the probability of an obvious edge
    alone the line connecting the two pixels:
    \begin{equation}
    W_{ij}=exp(-\max_{p\in\bar{ij}}{~mPb(p)}/\sigma)
    \end{equation}
%    $mPb(.)$ tells the probability of being an edge pixel,
    where $\bar{ij}$ is the line segment connecting $i$ and $j$ and $\sigma$ is a constant, the $mPb(p)$ is
    the boundary strength defined at pixel $p$ by maximizing the oriented contour signal $mPb(p, \theta)$ at multiple orientations $\theta$ :
    \begin{equation}
    mPb(p) = \max_{\theta}~{mPb(p, \theta)}
    \end{equation}
    The oriented contour signal $mPb(p, \theta)$ is defined as a linear cobination of multiple local cuess at orientation $\theta$:
    \begin{equation}
    mPb(p, \theta) = \sum_{s} \sum_{i} \alpha_{i,s}G_{i, \sigma(i,s)}(p, \theta)
    \end{equation}
    where $G_{i, \sigma(i,s)}(p, \theta)$ measures the $\chi^2$ distance at feature channel $i$ (brightness, color a, color b, texture) between the histograms of the two halves of a disc of radius $\sigma(i, s)$ divided at angle $\theta$, and $\alpha_{i,s}$ is the combination weight by gradient ascent on the F-measure using the training images and corresponding ground-truth. Moreover, the affinity matrix can also be learned by using the recent multi-task low-rank representation algorithm presented in~\cite{ChengLWHY11}. 
    \vspace{3pt}
    
    The segmentation is achieved by recursively bi-partitioning
    the graph using the first nonzero eigenvalue's eigenvector~\cite{ShiM00}
    or spectral clustering of a set of eigenvectors~\cite{NgJW01}. For
    the computational efficiency purpose, spectral clustering requires the affinity matrix
    to be sparse which limits its applications. Recent work of Cour et al.~\cite{CourBS05} solves
    this limitation by defining the affinity matrix at multiple scale and then setting up
    cross-scale constraints which achieve better result. In addition, Arbelaez et al.~\cite{ArbelaezMFM09} convolve eigenvectors with Gaussian directional derivatives at multiple orientations $\theta$ to obtain oriented spectral contours responses at each pixel $p$:
    \begin{equation}
    sPb(p, ~\theta) =\sum\limits_{k=1}^n \frac{1}{\sqrt{\lambda_k}}\cdot\triangledown_{\theta}\textbf{v}_k(p)
    \end{equation}
    Since the signal $mPb$ and $sPb$ carries different contour information, Arbelaez et al.~\cite{ArbelaezMFM09} proposed to combine them to globalized the contour information $gPb$:
    \begin{equation}\label{eq:gpb}
    gPb(p, \theta) = \beta\cdot mPb(p, \theta) + \gamma\cdot sPb(p, \theta)
    \end{equation}
    where the combination weights $\beta$ and $\gamma$ are also learned by gradient ascent on the F-measure using ground truth, which achieved the state-of-the-art contour detection result.

\textbf{Edge Based Region Merging}: Such methods
	\cite{Brice1970}\cite{JainTLB04}
	start from pixels or super-pixels, and then two adjacent regions are
	merged based on the metrics which can reflect their
	similarities/dissimilarities such as boundary length,
	edge strength or color difference. Recently,
	Arbelaez \emph{et al.}~\cite{ArbelaezMFM11}
	proposed the gPb-OWT-UCM method to transform
	a set of contours,
	which are generated from the Normalized Cut framework (to be introduced later),
	into a nested partition of the image.
	The method first generate a probability edge
	map $gPb$ (see Eq.\ref{eq:gpb}) which delineates the salient contours.
	Then, it performs watershed over the topological space defined
	by the $gPb$ to form the finest level segmentation.
	Finally, the edges between regions are sorted and merged in an
	ascending order which forms the ultrametric contour map $UCM$.
	Thresholding $UCM$ at a scale $\lambda$ forms the final segmentation.

%
%    The segmentation is achieved by recursively bi-partitioning
%    the graph using the first nonzero eigenvalue's eigenvector~\cite{ShiM00}
%    or spectral clustering of a set of eigenvectors~\cite{NgJW01}. For
%    the purpose of computational efficiency, spectral clustering requires the affinity matrix
%    to be sparse which limits its applications. Recent work of Cour \emph{et al.}~\cite{CourBS05} solves
%    this limitation by defining the affinity matrix at multiple scales and then setting up
%    cross-scale constraints which lead to better results.

\subsection{Continuous methods} \label{sec:bottom-up-ac}
Variational techniques~\cite{Osher88, Chan2001, BressonEVTO07,
PockCCB09, YuanBTB10} have also been used in segmenting images
into regions, which treat an image as a continuous surface instead
of a fixed discrete grid and can produce visually more pleasing
results.

\textbf{Mumford-Shah Model}: The Mumford-Shah (MS) model
partitions an image by minimizing the functional which encourages
homogeneity within each region as well as sharp piecewise regular
boundaries. The MS functional is defined as
\begin{equation}\label{mumford-shah}
    F_{MS}(I, C) = \int_{\omega}|I-I_0|^2 dx + \mu \int_{\omega \backslash C}|\nabla I|^2 dx + \nu H^{N-1}(C)
\end{equation}
for any observed image $I_0$ and any positive parameters
$\mu,\nu$, where $I$ corresponds to a piecewise smooth
approximation of $I_0$, $C$ represents the boundary contours of
$I$ and its length is given by Hausdorff measure $H^{N-1}(C)$.
    The first term of \eqref{mumford-shah} is a fidelity term with respect to the given data $I_0$,
    the second term regularizes the function $I$ to
    be smooth inside the region $\omega \backslash C$ and the last term imposes a regularization constraint on the
    discontinuity set $C$ to be smooth.

Since minimizing the Mumford-Shah model is not easy, many variants
have been proposed to approximate the
functional~\cite{AmbrosioTOR90, Braides97, Braides98a,
Chambolle95}. Among them, Vese-Chan\cite{Vese2002} proposed to
approximate the term $H^{N-1}(C)$ by the lengths of region
contours, which provides the model of \textit{active contour
without edges}. By assuming the region is piecewise constant, the
model is further simplified to the continuous Potts model, which
has convexified solvers~\cite{PockCCB09, YuanBTB10}.

\textbf{Active Contour / Snake Model}~\cite{Kass1988}: This type
of models detects objects by deforming a snake/contour curve $C$
towards the sharp image edges. The evolution of parametric curve
$C(p)=(x(p), y(p)), p\in\{0, 1\}$ is driven by minimizing the
functional:
    \begin{equation}\label{active-contour}
    %\begin{split}
    F(C) = \alpha\int_{0}^{1} |\frac{\partial C(p)}{\partial p}|^2dp %\\
     + \beta\int_{0}^{1} |\frac{\partial^2 C}{\partial p^2}|^2dp %\\
     + \lambda \int_{0}^{1} f^2(I_0(C))dp
    %\end{split}
    \end{equation}
where the first two terms enforce smoothness constraints by making
the snake act as a membrane and a thin plate correspondingly, and
the sum of the first two terms makes the \textit{internal energy}.
The third term, called \textit{external energy}, attracts the
curve toward the object boundaries by using the edge detecting
function
    \begin{equation}\label{edge detection}
    f(I_0) = \frac{1}{1+\gamma|\nabla(I_0*G_{\sigma})|^2}
    \end{equation}
where $\gamma$ is an arbitrary positive constant and
$I_0*G_{\sigma}$ is the Gaussian smoothed version of $I_0$. The
energy function is non-convex and sensitive to initialization. To
overcome the limitation, Osher \emph{et al.}~\cite{Osher88}
proposed the \textit{level set method}, which implicitly
represents curve $C$ by a higher dimension $\psi$, called the
level set function. Moreover, Bresson \emph{et
al.}~\cite{BressonEVTO07} proposed the convex relaxed active contour
model which can achieve desirable global optimal.

The bottom-up methods can also be classified into another two
categories: the
ones~\cite{FelzenszwalbH04}\cite{ShiM00}\cite{ArbelaezMFM11} which
attempt to produce regions likely belonging to objects and the
ones which tend to produce
over-segmentation~\cite{VincentS91}\cite{ComaniciuM02} (to be
introduced in Section~\ref{sec:superpixel}). For methods of the
first category, obtaining object regions is extremely challenging
as the bottom-up methods only use low-level features. Recently,
Zhu \emph{et al.}~\cite{ZhuZCM13} proposed to combine hand-crafted
low-level features that can reflect global image statistics (such
as Gaussian Mixture Model, Geodesic Distance and Eigenvectors)
with the convexified continuous Potts model to capture high-level
structures, which achieves some promising results.

\section{Superpixel}
\label{sec:superpixel}
%As it is still difficult to segment an image into object regions reliably by
%using existing popular low level cues (intensity, color, texture and curvature \emph{et al.})
%and models,
Superpixel methods aim to over-segment an image into homogeneous
regions which are smaller than object or parts. In the seminal
work of Ren and Malik~\cite{RenM03}, they argue and justify that
superpixel is more natural and efficient representation than pixel
because local cues extracted at pixel are ambiguous and sensitive
to noise. Superpixel has a few desirable properties: \bi
    \item \textbf{It is perceptually more meaningful than pixel.}
    The superpixel produced by state-of-the-art
    algorithms is nearly perceptually consistent in terms of color, texture, etc, and
    most structures can be well preserved. Besides, superpixel also conveys some shape cues which
    is difficult to capture at pixel level.
\item \textbf{It helps reduce model complexity and improve
efficiency and accuracy.} Existing pixel-based methods need to
deal with millions of pixels and their parameters, where training
and inference in such a big system pose great
    challenges to current solvers.  On the contrary, using superpixels to represent the image can greatly reduce
    the number of parameters and alleviate computation cost.  Meanwhile, by exploiting the
    large spatial support of superpixel, more discriminative features such as color or texture
    histogram can be extracted. Last but not least, superpixel makes longer range information
    propagation possible, which allows existing solvers to exploit richer information than those using pixel.
\ei

There are different paradigms to produce superpixels:
\begin{itemize}
\item Some existing bottom-up methods can be directly adapted to
over-segmentation scenario
           by tuning the parameters, e.g. Watersheds,
            Normalized Cut (by increasing cluster number),
            Graph Based Merging (by controlling the regions size)
            and Mean-Shift/Quick-Shift (by tuning the kernel size or changing mode drifting style).
\item Some recent methods produce much faster superpixel
segmentation by changing optimization scope from the whole image
to local non-overlap initial regions, and then adjusting the
region boundaries to snap to salient object contours.
TurboPixel~\cite{LevinshteinSKFDS09} deforms the initial spatial
grid to compact and regular regions by using geometric flow which
is directed by local gradients. Wang
\emph{et al.}~\cite{WangZGWZ13} also adapted geodesic flows by
computing geodesic distance among pixels to produce adaptive
superpixels, which have higher density in high intensity or color
variation regions while having larger superpixels at
structure-less regions. Veskler \emph{et al.}~\cite{VekslerBM10}
proposed to place overlapping patches at the image, and then
assigned each pixel by inferring the MAP solution using
graph-cuts. Zhang \emph{et al.}~\cite{ZhangHMB11} further studied
in this direction by using a pseudo-boolean optimization which
achieves faster speed. Achanta \emph{et al.}~\cite{AchantaSSLFS12}
introduced the SLIC algorithm which greatly improves the
superpixel efficiency. SLIC starts from the initial regular grid
of superpixels, grows superpixels by estimating each pixel's
distance to its cluster center localized nearby, and then updates
the cluster centers, which is essentially a localized K-means.
SLIC can produce superpixels at 5Hz without GPU optimization.
\item There are also some new formulations for over-segmentation.
Liu \emph{et al.}~\cite{Liu2014} recently proposed a new graph
based method, which can maximize the entropy rate of the cuts in
the graph with a balance term for compact representation. Although
it outperforms many methods in terms of boundary recall measure,
it takes about 2.5s to segment an image of size 480x320. Van den
Berge~\emph{et al.}~\cite{BerghBRCG12} proposed the fastest
superpixel method-SEED, which can run at 30Hz. SEED uses
multi-resolution image grids as initial regions. For each image
grid, they define a color histogram based entropy term and an
optional boundary term. Instead of using EM as in SLIC, which
needs to repeatedly compute distances,  SEEDs uses Hill-Climbing
to move coarser-resolution grids, and then refines region boundary
using finer-resolution grids. In this way, SEEDs can achieve real
time superpixel segmentation at 30Hz.
\end{itemize}

%----------------------------------------------------------------------
  \begin{figure}
    \begin{center}
    \includegraphics[width = 1\columnwidth]{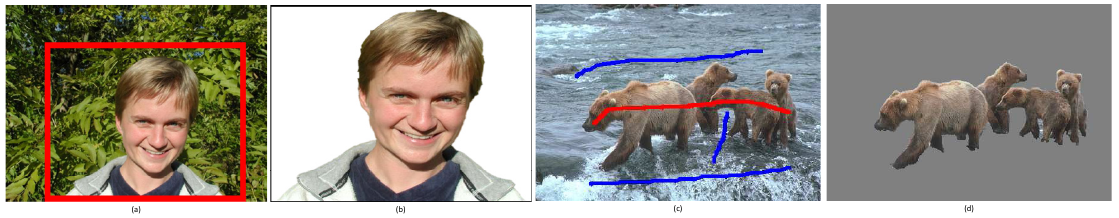}
    \end{center}
    \caption
    {\label{fig:interseg} Examples of interactive image segmentation
        using bounding box (a $\sim$ b)~\cite{RotherKB04} or scribbles (c $\sim$
        d)~\cite{YangCZL10}.
    }
  \end{figure}
  %------------------------

\section{Interactive methods}
\label{sec:interactive} Image segmentation is expected to produce
regions matching human perception. Without any prior assumption,
it is difficult for bottom-up methods to produce object regions.
For some specific areas such as image editing and medical image
analysis, which require precise object localization and
segmentation for subsequent applications (e.g. changing background
or organ reconstruction), prior knowledge or constraints (e.g.
color distribution, contour enclosure or texture distribution)
directly obtained from a small amount of user inputs can be of
great help to produce accurate object segmentation. Such type of
segmentation methods with user inputs are termed as interactive
image segmentation methods. There are already some
surveys~\cite{McGuinnessO10, Yi2012, He2013} on interactive
segmentation techniques, and thus this paper will act as a
supplementary to them, where we discuss recent influential
literature not covered by the previous surveys. In addition, to
make the manuscript self-contained, we will also give a brief
review of some classical techniques.

In general, an interactive segmentation method has the following
pipeline: 1) user provides initial input; 2) then segmentation
algorithm produces segmentation results; 3) based on the results,
user provides further constraints, and then go back to step 2. The
process repeats until the results are satisfactory.  A good
interactive segmentation method should meet the following
criteria: 1) offering an intuitive interface for the user to
express various constraints; 2) producing fast and accurate result
with as little user manipulation as possible; 3) allowing user to
make additional adjustments to further refine the result.

Existing interactive methods can be classified according to the
difference of the user interface, which is the bridge for the user
to convey his prior knowledge (see Fig.~\ref{fig:interseg}).
Existing popular interfaces include bounding
box~\cite{RotherKB04}, polygon (the object of interest is within
the provided regions), contour~\cite{KassWT88, Mortensen1992} (the
object boundary should follow the provided direction), and
scribbles~\cite{BoykovJ01,YangCZL10}
 (the object should follow similar color distributions). According to the methodology, the existing interactive methods
can be roughly classified into two groups: (i) contour based
methods and (ii) label propagation based methods~\cite{He2013}.

\subsection{Contour based interactive methods}
Contour based methods are one type of the earliest interactive
segmentation methods. In a typical contour based interactive
method, the user first places a contour close to the object
boundary, and then the contour will evolve to snap to the nearby
salient object boundary. One of the key components for contour
based methods is the edge detection function, which can be based
on first-order image statistics (such as operators of Sobel,
Canny, Prewitt, etc), or more robust second-order statistics (such
as operators of Laplacian, LoG, etc).

For instance, the \textit{Live-Wire / Intelligent
Scissors} method~\cite{Mortensen1992, MortensenB95} starts contour
evolving by building a weighted graph on the image, where each node in the graph corresponds to a pixel, and directed edges are formed
around pixels with their closest four neighbors or eight neighbors. The local
cost of each direct edge is the weighted sum of Laplacian
zero-cross, gradient magnitude and gradient direction. Then given
the seed locations, the shortest path from a seed point $p$ to a certain
seed point $s$ is found by using Dijikstra's method.
Essentially, Live-Wire minimizes a local energy function. On the
other hand, the active contour method introduced in Section~\ref{sec:bottom-up-ac}
deforms the contour by using a global energy function, which
consists of a regional term and a boundary term, to overcome
some ambiguous local optimum. Nguyen~\emph{et al.}~\cite{AnhCZZ12}
recently employed the convex active contour model to refine the object
boundary produced by other interactive segmentation methods and
achieved satisfactory segmentation results. Liu \emph{et
al.}~\cite{LiuY12} proposed to use the level set
function~\cite{Osher88} to track the zero level set of the posterior probabilistic
mask learned from the user provided bounding box, which can capture
objects with complex topology and fragmented appearance
such as tree leaves.

\subsection{Label propagation based methods}
Label Propagation methods are more popular in literature. The basic idea of label
propagation is to start from user-provided initial input marks, and then propagate the labels
using either global optimization (such as
GraphCut~\cite{BoykovJ01} or RandomWalk~\cite{Grady06}) or local
optimization, where the global optimization methods are more widely used
due to the existence of fast solvers.

\textbf{GraphCut and its decedents}~\cite{BoykovJ01,
RotherKB04, LiSTS04} model the pixel labeling problem in Markov
Random Field. Their energy functions can be generally expressed as
    \begin{equation}\label{eq:graphcut}
       E (X) = \sum_{x_i \in X} E_u(x_i) + \sum_{x_i, x_j \in \cal{N}} E_p(x_i, x_j)
    \end{equation}
where $X = {x_1, x_2,..., x_n}$ is the set of random variables
defined at each pixel, which can take either foreground label $1$ or
background label $0$; $\cal{N}$ defines a neighborhood system,
which is typically 4-neighbor or 8-neighbor. The first term in~\eqref{eq:graphcut} is the unary potential, and the second term is the pairwise
potential. In~\cite{BoykovJ01}, the unary potential is evaluated by using
an intensity histogram. Later in GrabCut~\cite{RotherKB04}, the unary potential is derived from the two Gaussian Mixture Models (GMMs) for background and foreground regions
respectively, and then a hard constraint is imposed on the regions
outside the bounding box such that the labels in the background
region remain constant while the regions within the box are
updated to capture the object. Li \emph{et al.}~\cite{LiSTS04}
further proposed the LazySnap method, which extends GrabCut by using superpixels and including an interface to
allow user to adjust the result at the low-contrast and weak
object boundaries. One limitation of GrabCut is that it favors
short boundary due to the energy function design. Recently Kohli
\emph{et al.}~\cite{KohliOJ13} proposed to use a conditional
random field with multiple-layered hidden units to encode boundary
preserving higher order potential, which has efficient solvers and
therefore can capture thin details that are often neglected by
classical MRF model.

Since MRF/CRF model provides a unified framework to combine multiple
information, various methods have been proposed to incorporate
prior knowledge:
\bi
\item Geodesic prior: One typical assumption on objects
is that they are compactly clustered in spatial space, instead of
distributing around. Such spatial constraint can be constructed by
exploiting the geodesic distance to foreground and background
seeds. Unlike the Euclidean distance which directly measures two-point distance in spatial space, the geodesic distance measures
the lowest cost path between them. The weight is set to the
gradient of the likelihood of pixels belonging to the foreground. Then
the geodesic distance can be incorporated as a kind of data term in
the energy function as in~\cite{PriceMC10a}. Later, Bai \emph{et
al.}~\cite{BaiS07} further extended the solution to soft matting
problem.  Zhu \emph{et al.}~\cite{ZhuZCM13} also incorporated the
Bai's geodesic distance as one type of object potential to produce
bottom-up object segmentation, which achieves improved results.

\item Convex  prior: Another common assumption is that most
objects are convex. Such prior can be expressed as a kind of
labeling constraints. For example, a star shape is defined with respect to a
center point $c$. An object has a star shape if for any point $p$
inside the object, all points on the straight line between the
center $c$ and $p$ also lie inside the object. Veskler \emph{et
al.}~\cite{Veksler08} formulated such constraint by penalizing
different labeling on the same line, which such formulation can only
work with single convex center. Gulshan \emph{et
al.}~\cite{GulshanRCBZ10} proposed to use geodesic distance
transform~\cite{CriminisiSB08} to compute the geodesic convexity
from each pixel to the star center, which works on objects with
multiple start centers. Other similar connectivity constraint has
also been studied in~\cite{VicenteKR08}.
\ei

\textbf{RandomWalk}: The RandomWalk model~\cite{Grady06} provides
another pixel labeling framework, which has been applied to many
computer vision problems, such as segmentation, denoising and
image matching. With notations similar to those for GraphCut
in~\eqref{eq:graphcut}, RandomWalk starts from building an
undirected graph $G=(V, E)$, where $V$ is the set of vertices
defined at each pixel and $E$ is the set of edges which
incorporate the pairwise weight $W_{ij}$ to reflect the
probability of a random walker to jump between two nodes $i$ and
$j$. The \textit{degree} of a vertex is defined as $d_i = \sum_j
W_{ij}$.

Given the weighted graph, a set of user scribbled nodes $V_m$ and
a set of unmarked nodes $V_u$, such that $V_m \cup V_u = V$ and
$V_m \cap\ V_u = \varPhi$, the RandomWalk approach is to assign
each node $i \in V_u$ a probability $x_i$ that a random walker
starting from that node first reaches a marked node. The final
segmentation is formed by thresholding $x_i$.

The entire set of node probability $x$ can be obtained by
minimizing
    \begin{equation}\label{random walk}
    E (x) = x^{T}Lx
    \end{equation}
where $L$ represents the combinatorial Laplacian matrix defined as
\begin{equation}
L_{ij} =
\begin{cases}
d_{i}  & \text{if}~  i=j \\
-w_{ij} & \text{if $i\neq j$ and $i$ and $j$ are adjacent nodes}~ \\
0 & \text{otherwise.}
\end{cases}
\end{equation}
By partitioning the matrix $L$ into marked and unmarked blocks as
\begin{equation}
\left[
\begin{array}{cc}
L_m & B\\
B^T & L_u
\end{array}
\right]
\end{equation}
and defining a $|V_m|\times 1$ indicator vector $f$ as
\begin{equation}
f_j =
\begin{cases}
1  & \text{if}~  j~\text{is labeled as foreground} \\
0 & otherwise,
\end{cases}
\end{equation}
the minization of~\eqref{random walk} with respect to $x_u$ results in
\begin{equation}
L_u x_u = -B^T f
\end{equation}
where only a sparse linear system needs to be solved. Yang \emph{et al.}~\cite{YangCZL10} further proposed a constrained RandomWalk that is able to incorporate different types of user inputs as additional constraints such as hard and soft constraints to provide more flexibility for users.

\textbf{Local optimization based methods}: Many label propagation methods use specific solvers such as graph cut and
belief propagation to get the most likely solution. However, such global optimization type of
solvers are often slow and do not scale with image size. Hoshi \emph{et al.}~\cite{HosniRBRG13} proved that by filtering
the cost volume using fast edge preserving filters (such as Joint
Bilateral Filter~\cite{DurandD02}, Guidede Filter~\cite{He0T13},
Cross-map filter~\cite{LuSMLD12}, etc) and then using
Winner-Takes-All label selection to take the most
likely labels, they can achieve comparable or better results than global
optimized models. More importantly, such cost-volume filtering approaches can achieve
real time performance, e.g. 2.85 ms to filter an 1 Mpix image on a Core 2
Quad 2.4GHZ desktop. Recently, Crimisi \emph{et
al.}~\cite{CriminisiSB08} proposed to use geodesic distance
transform to filter the cost volume, which can produce results comparable
to the global optimization methods and can better capture
the edges at the weak boundaries.

There are also some simple region
growing based methods~\cite{Adams1994, NingZZW10}, which start from user-drawn seeded regions, and then iteratively merge the remaining groups according to
similarities. One limitation of such methods is that different
merging orders can produce different results.

\section{Object Proposals}
Automatically and precisely segmenting out objects from an image
is still an unsolved problem. Instead of searching for
deterministic object segmentation, recent research on object
proposals relaxes the object segmentation problem by looking for a
pool of regions that have high probability to cover the objects by some of the proposals. This type of methods leverages high
level concept of ``object'' or ``thing'' to separate object
regions from ``stuff'', where an ``object'' tends to have clear
size and shape (e.g. pedestrian, car), as opposed to
``stuff''(e.g. sky or grass) which tends to be homogeneous or with
recurring patterns of fine-scale structure.

\textbf{Class-specific object proposals}: One approach to
incorporate the object notion is through the use of class-specific
object detectors. Such object detectors can be any bounding box
detector, e.g. the famous Deformable Part Model
(DPM)~\cite{FelzenszwalbGM10} or Poselets~\cite{BourdevMBM10}. A
few works have been proposed to combine object detection with
segmentation. For example, Larlus and Jurie~\cite{LarlusJ08}
obtained the object segmentation by refining the bounding box
using CRF. Gu \emph{et al.}~\cite{GuLAM09} proposed to use
hierarchical regions for object detection, instead of bounding
boxes. However, class-specific object segmentation methods can
only be applied to a limited number of object classes, and cannot
handle large number of object classes, e.g.
ImageNet~\cite{DengDSLL009}, which has thousands of classes.

%----------------------------------------------------------------------
 \begin{figure}
    \begin{center}
    \includegraphics[width = 1\columnwidth]{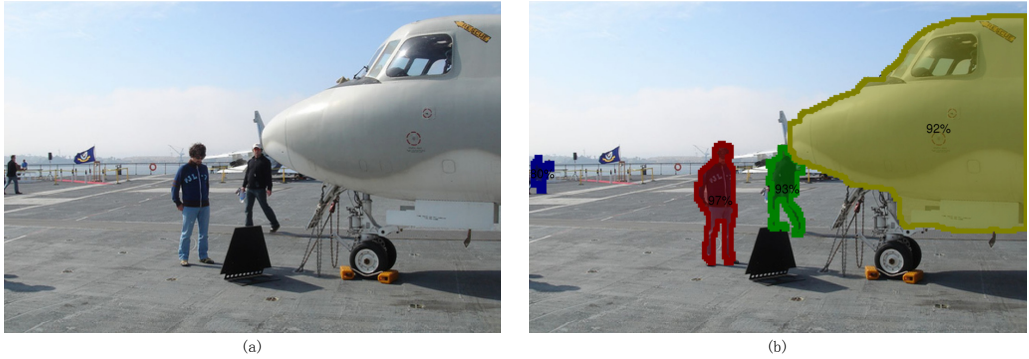}
    \caption{\label{fig:objectness} An example of
    class-independent object segmentation~\cite{CarreiraS12}.}
\end{center}
 \end{figure}
 %------------------------

\textbf{Class-independent object proposals}:
Inspired by the objectness window work
of Alexi \emph{et al.}~\cite{AlexeDF12}, methods of class-independent region proposals directly attempt to produce general
object regions. The underlying rationale for class-independent proposals to work is that the
object-of-interest is typically distinct from background in certain
appearance or geometry cues. Hence, in some sense general object
proposal problem is correlated with the popular salient object
detection problem. A more complete survey of
state-of-the-art objectness window methods and salient object detection methods can
be found in~\cite{Hosang14} and~\cite{BorjiCJL14}, respectively. Here we
focus on the region proposal methods.

One group of class-independent object proposal methods is to first use bottom-up methods to generate
regions proposals, and then a pre-trained classifier is applied to
rank proposals according to the region features. The representative works include CPMC~\cite{CarreiraS12} and Category
Independent Object Proposal~\cite{EndresH14}, which extend GrabCut to this scenario by using different seed
sampling techniques (see
Fig.~\ref{fig:objectness} for an example). In particular, CPMC applies sequential GrabCuts by using the regular grid as foreground seeds and image frame boundary as background seeds to train the Gaussian mixture models.
Category Independent Object Proposal samples foreground
seeds from occlusion boundaries inferred by~\cite{HoiemEH11}. Then
the generated regions proposals are fed to random forests
classifier or regressor, which is trained with the the features
(such as convexity, color contrast, etc) of the ground truth
object regions, for re-ranking. One bottleneck of
such appearance based object proposal methods is that re-computing
GrabCut's appearance model implicitly incorporates the
re-computation of some distance, which makes such
methods hard to speed up. A recent
work in~\cite{HumayunLR14} pre-computes a
graph which can be used for parametric min-cuts over different
seeds, and then dynamic graph cut~\cite{KohliRBT08} is applied to
accelerate the speed. Instead of computing expensive GrabCut, another recent work called Geodesic Object Proposal
(GOP)~\cite{KrahenbuhlK14} exploits the fast
geodesic distance transform for object proposals.
GOP uses superpixels as its
atomic units, chooses the first seed location
as the one whose geodesic distance is the smallest to all other
superixels, and then places next seeds far to the existing seeds.
The authors also proposed to use RankSVM to learn to place seeds, but the performance improvement is not significant.

Another bottom-up way to generate region proposals is to use the edge based region merging method, gPb-OWT-UCM, described in Section~\ref{sec:bottom-up}. Then, the region hierarchies are
filtered using the classifiers of CPMC. Due to its reliance on the
high accuracy contour map, such method can achieve higher accuracy
than GrabCut based methods. However, such method is limited
by the performance of contour detectors, whose shortcomings on speed and accuracy
have been greatly improved by some recently introduced learning based
methods such as Structured Forest~\cite{DollarZ13} or the
multi-resolution eigen-solvers. The later solver has been applied
in an improved version of gPb-OWT-UCM (MCG)~\cite{ArbelaezPBMM14}
which simultaneously considers the region combinatorial space.

Different from the above class-independent object proposal methods, which use single strategy to generate object
regions, another group of methods apply multiple hand-crafted strategies to
produce diversified solutions during the process of atomic unit
generation and region proposal generation, which we call
\textit{diversified region proposal} methods. This type of methods typically just produce diversified region proposals, but do not train classifier to rank the proposals. For example, SelectiveSearch~\cite{SandeUGS11} generates
region trees from superpixels to capture objects at multiple
scales by using different merging similarity metrics (such as RGB,
Intensity, Texture, etc). To increase the degree of
diversification, SeleciveSearch also applies different parameters
to generate initial atomic superpixels. After different
segmentation trees are generated, the detection starts from the
regions in higher hierarchies. Manen \emph{et
al.}~\cite{ManenGG13} proposed a similar method which
exploits merging randomized trees with learned similarity metrics.

\section{Semantic Image Parsing}
\label{sec:semantic seg} 
\begin{figure}
    \begin{center}
    \includegraphics[width = 1\columnwidth]{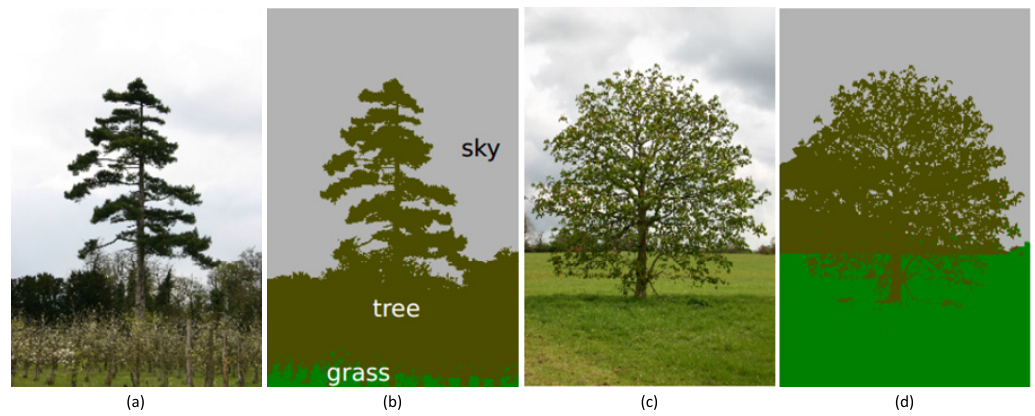}
    \end{center}
    \caption
    {\label{fig:semanseg} An example of semantic segmentation~\cite{KrahenbuhlK11} which is trained using pictures with human labeled ground truth such as (b) to segment the test image in (c) and produce the final segmentation in (d).
    }
  \end{figure}

Semantic image parsing aims to break an
image into non-overlapped regions which correspond to predefined
semantic classes (e.g. car, grass, sheep, etc), as shown in
Fig.~\ref{fig:semanseg} . The popularity of semantic parsing since
early 2000s is deeply rooted in the success of some specific
computer vision tasks such as face/object detection and tracking,
camera pose estimation, multiple view 3D reconstruction and fast
conditional random field solvers. The ultimate goal of semantic
parsing is to equip the computer with the holistic ability to
understand the visual world around us. Although also depending on
the given information, high-level learned representations make it
different from the interactive methods. The learned models can be
used to predict similar regions in new images. This type of
approaches is also different from the object region proposal in the
sense that it aims to parse an image as a whole into the ``thing''
and ``stuff'' classes, instead of just producing possible
``thing'' candidates.

%----------------------------------------------------------------------
%------------------------

Most state-of-the-art image parsing systems are formulated as the
problem of finding the most probable labeling on a Markov random
field (MRF) or conditional random field (CRF). CRF provides a
principled probabilistic framework to model complex interactions
between output variables and observed features. Thanks to the
ability to factorize the probability distribution over different
labeling of the random variables, CRF allows for compact
representations and efficient inference. The CRF model defines a
Gibbs distribution of the output labeling \textbf{y} conditioned
on observed features \textbf{x} via an energy function
$E(\textbf{y}; \textbf{x})$:
\begin{equation}\label{crf}
    P(\textbf{y}|\textbf{x}) = \frac{1}{Z(\textbf{x})}exp\{-E(\textbf{y};
    \textbf{x})\},
\end{equation}
where $\textbf{y} =(y_1, y_2, …., y_n)$ is a vector of random
variables $y_i$ ($n$ is the number of pixels or regions) defined
on each node $i$ (pixel or superpixel), which  takes  a label from
a predefined label set $L$ given the observed features
$\textbf{x}$. $Z(\textbf{x})$ here is called partition function
which ensures the distribution is properly normalized and summed
to one. Computing the partition function is intractable due to the
sum of exponential functions. On the other hand, such computation
is not necessary given the task is to infer the most likely
labeling. Maximizing a posterior of~\eqref{crf} is equivalent to
minimize the energy function $E(\textbf{y}; \textbf{x})$. A common
model for pixel labeling involves a unary potential $\psi^u(y_i;
x)$  which is associated with each pixel, and a pairwise potential
$\psi^p(y_i, y_j)$ which is associated with a pair of neighborhood
pixels:
\begin{equation}\label{energyfunc2}
    E(\textbf{y}; \textbf{x}) = \sum_{i=1}^n \psi^u(y_i; \textbf{x}) + \sum_{v_i, v_j \in \cal{N}} \psi_{ij}^p(y_i, y_j;
    \textbf{x}).
\end{equation}

Given the energy function, semantic image parsing usually follows
the following pipelines: 1) Extract features from a patch centered
on each pixel; 2) With the extracted features and the ground truth
labels, an appearance model is trained to produce a compatible
score for each training sample; 3) The trained classifier is
applied on the test image's pixel-wise features, and the output is
used as the unary term; 4) The pairwise term of the CRF is defined
over a 4 or 8-connected neighborhood for each pixel; 5) Perform
maximum a posterior (MAP) inference on the graph. Following this
common pipeline, there are different variants in different
aspects:
\begin{itemize}
\item \textbf{Features}: The commonly used features are bottom-up
pixel-level features such as color or texton. He \emph{et
al.}~\cite{HeZC04} proposed to incorporate the region and image
level features. Shotton \emph{et al.}~\cite{ShottonJC08} proposed
to use spatial layout filters to represent the local information
corresponding to different classes, which was later adapted to
random forest framework for real-time parsing~\cite{ShottonWRC09}.
Recently, deep convolutional neural network learned
features~\cite{Farabet2013} have also been applied to replace the
hand-crafted features, which achieves promising
performance.
\item \textbf{Spatial support}:  The spatial support in step 1 can
be adapted to superpixels which conform to image internal
structures and make feature extraction less susceptible to noise.
Also by exploiting superpixels, the complexity of the model is
greatly reduces from millions of variables to only hundreds or
thousands. Hoiem \emph{et al.}~\cite{HoiemEH07} used multiple
segmentations to find out the most feasible configuration. Tighe
\emph{et al.}~\cite{Tighe2013} used superpixels to retrieve
similar superpixels in the training set to generate unary term. To
handle multiple superpixel hypotheses, Ladicky \emph{et
al.}~\cite{LadickyRKT09} proposed the robust higher order
potential, which enforces the labeling consistency between the
superpixels and their underlying pixels.
\item \textbf{Context}: The context (such as boat in the water,
car on the road) has emerged as another important factor beyond
the basic smoothness assumption of the CRF model. Basic context
model is implicitly captured by the unary potential, e.g. the
pixels with green colors are more likely to be grass class.
Recently, more sophisticated class co-occurrence information has
been incorporated in the model. Rabinovich \emph{et
al.}~\cite{RabinovichVGWB07} learned label co-occurence statistic
in the training set and then incorporated it into CRF as
additional potential. Later the systems using multiple forms of
context based on co-occurence, spatilal adjacency and appearance
have been proposed in~\cite{GalleguillosRB08, GalleguillosMBL10,
Tighe2013}. Ladicky \emph{et al.}~\cite{LadickyRKT10} proposed an
efficient method to incorporate global context, which penalizes
unlikely pairs of labels to be assigned anywhere in the image by
introducing one additional variable in the GraphCut model.
\item \textbf{Top-down and bottom-up combination}:  The
combination of top-down and bottom-up information has also been
considered in recent scene parsing works. The bottom-up
information is better at capturing stuff class which is
homogeneous. On the other hand, object detectors are good at
capturing thing class. Therefore, their combination helps develop
a more holistic scene understanding system. There are some recent
studies incorporating the sliding window detectors such as
Deformable Part Model~\cite{FelzenszwalbMR08} or
Poselet~\cite{BourdevMBM10}. Specifically, Ladicky \emph{et
al.}~\cite{LadickySART10} proposed the higher order robust
potential based on detectors which use GrabCut to generate the
shape mask to compete with bottom up cues. Floros \emph{et
al.}~\cite{FlorosRL11} instead infered the shape mask from the
Implicit Shape Model top-down segmentation
system~\cite{LeibeLS08}. Arbelaez \emph{et
al.}~\cite{ArbelaezHGGBM12} used the Poselet detector to segment
articulated objects. Guo and Hoiem~\cite{GuoH12} chose to use
Auto-Context~\cite{Tu08} to incorporate the detector response in
their system. More recently, Tighe \emph{et al.}~\cite{TigheL13}
proposed to transfer the mask of training images to test images as
the shape potential by using trained exemplar SVM model and
achieved state-of-the-art scene parsing results.
\item \textbf{Inference}: To optimize the energy function, various
techniques can be applied, such as GraphCut, Belief-Propagation or
Primal-Dual methods, etc. A complete review of recent inference
methods can be found in~\cite{KappesAHSNBKKLKR13}. Original CRF or
MRF models are usually limited to 4-neighbor or 8-neighbor.
Recently, the fully connected graphical model which connects all
pixels has also become popular due to the availability of
efficient approximation of the time-costly message-passing step
via fast image filtering~\cite{KrahenbuhlK11}, with the
requirement that the pairwise term should be a mixture of Gaussian
kernels. Vineet \emph{et al.}~\cite{VineetWT14} introduced the
higher order term to the fully connected CRF framework, which
generalizes its application.
\item \textbf{Data Driven}: The current pipeline needs pre-trained
classifier, which is quite restrictive when new classes are
included in the database. Recently some researchers have advocated
for non-parametric, data-driven approach for open-universe
datasets. Such approaches avoid training by retrieving similar
training images from the database for segmenting the new image.
Liu \emph{et al.}~\cite{LiuYT11a} proposed to use
SIFT-flow~\cite{LiuYT11} to transfer masks from train images. On
the other hand, Tighe \emph{et al.}~\cite{Tighe2013} proposed to
retrieve nearest superpixel neighbor in training images and
achieved comparable performance to~\cite{LiuYT11a}.
\end{itemize}

  %----------------------------------------------------------------------
  \begin{figure*}
    \begin{center}
        \includegraphics[width = 1\columnwidth]{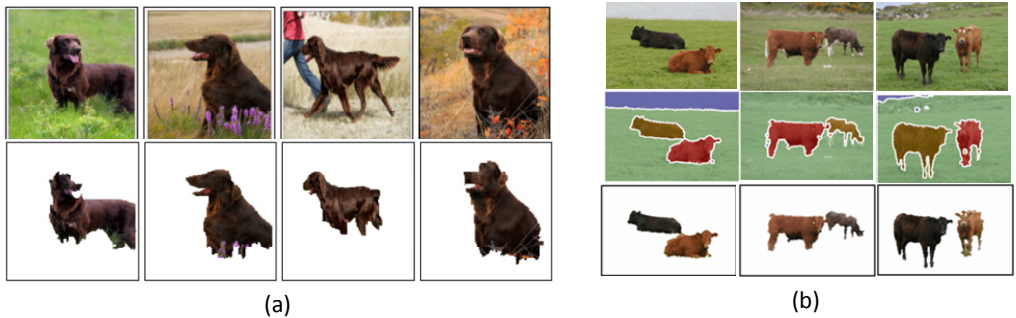}
    \end{center}
    \caption
    {\label{fig:coseg} (a) An example of simultaneously segmenting one common foreground object from a set of image~\cite{ChangLL11}. (b) An example of multi-class cosegmentation~\cite{KimXLK11}.
    }
  \end{figure*}
  %------------------------

\section{Image Cosegmentation} \label{sec:coseg}
Cosegmentation aims at extracting common objects from a set of
images (see Fig.~\ref{fig:coseg} for examples). It is essentially a
multiple-image segmentation problem, where the very weak
prior that the images contain the same objects is used for
automatic object segmentation. Since it does not need any pixel
level or image level object information, it is suitable
for large scale image dataset segmentation and many practical
applications, which attracts much attention recently. At the meanwhile, cosegmentation is challenging, which faces several major issues:
\begin{itemize}
\item The classical segmentation models are designed for a single image
while cosegmentation deals with multiple images. How to design
the cosegmentation model and minimize the model is critical
for cosegmentation research.
\item The common object extraction depends on the foreground
similarity measurement. But the foreground usually varies,
which makes the foreground similarity measurement difficult. Moreover, the
model minimization is highly related to the selection of similarity
measurement, which may result in extremely difficult optimization. Thus, how to measure the foreground similarities
is another important issue.
\item There are many practical applications with different
requirements such as large-scale image cosegmentation, video
cosegmentation and web image cosegmentation. Individual special
needs require a cosegmentation
model to be extendable to different scenarios, which is challenging.
\end{itemize}

Cosegmentation was first introduced by Rother \emph{et
al.}~\cite{RotherMBK06} in 2006. After that, many cosegmentation
methods have been proposed~\cite{KimX12, KimXLK11, JoulinBP12,
RubioSLP12, JoulinBP10, ChangLL11, MukherjeeSP11,MukherjeeSD09}.
The existing methods can be roughly classified into three
categories. The first one is to extend the existing single image
based segmentation models to solve the cosegmentation problem,
such as MRF cosegmentation, heat diffusion cosegmentation, RandomWalk based cosegmentation and active contour based cosegmentation.
The second one is to design new cosegmentation models, such as
formulating the cosegmentation as clustering problem, graph theory
based proposal selection problem, metric rank based
representation. The last one is to solve new emerging
cosegmentation needs, such as multiple class/foreground object
cosegmentation, large scale image cosegmentation and web image
cosegmentation.

\subsection{Cosegmentation by extending single-image segmentation models}
A straight-forward way for cosegmentation is to extend classical single image
based segmentation models. In
general, the extended models can be represented as
\begin{equation}\label{fn:cosegmodel}
E = E_s + E_g
\end{equation}
where $E_s$ is the single image segmentation term, which
guarantees the smoothness and the distinction
between foreground and background in each image, and $E_g$ is the
cosegmentation term, which focuses on evaluating the consistency
between the foregrounds among the images. Only the segmentation of
common objects can result in small values of both $E_s$ and
$E_g$. Thus, the cosegmentation is formulated as minimizing the
energy in~\eqref{fn:cosegmodel}.

Many classical segmentation models have been used to form $E_s$, e.g.
using MRF segmentation models~\cite{RotherMBK06, MukherjeeSD09} as
\begin{equation}\label{fn:cosegmodel2}
E_s^{MRF} = E_u^{MRF} + E_p^{MRF}
\end{equation}
where $E_u^{MRF}$ and
$E_p^{MRF}$ are the conventional unary potential and the pairwise potential. GraphCut algorithm
is widely used to minimize the energy
in~\eqref{fn:cosegmodel2}. The cosegmentation term $E_g$ is used
to evaluate the multiple foreground consistency, which is
introduced to guarantee the common object segmentation. However, it makes the minimization
of ~\eqref{fn:cosegmodel} difficult. Various cosegmentation terms
and their minimization methods have been carefully designed in MRF
based cosegmentation models.

In particular, Rother \emph{et al.}~\cite{RotherMBK06} evaluated the consistency
by $\ell_1$ norm, i.e., $E_g = \sum_z(|h_1(z) - h_2(z)|)$, where $h_1$
and $h_2$ are features of the two foregrounds, and $z$ is the dimension of
the feature. Adding the
$\ell_1$ evaluation makes the minimization quite challenging. An
approximation method called submodular-supermodular procedure has been
proposed to minimize the model by max-flow algorithm. Mukherjee
\emph{et al.}~\cite{MukherjeeSD09} replaced $\ell_1$ with squared $\ell_2$
evaluation, i.e. $E_g = \sum_z(|h_1(z) -
h_2(z)|)^2$. The $\ell_2$ has several advantages, such as
relaxing the minimization to LP problem and using Pseudo-Boolean
optimization method for minimization. But it is still
an approximation solution. In order to simplify the model
minimization, Hochbaum \emph{et al.}~\cite{HochbaumS09} used
reward strategy rather than penalty strategy. The rationale is
that given any foreground pixel $p$ in the first image, the
similar pixel $q$ in the second image will be rewarded as
foreground. The global term was formulated as $E_g = \sum_z
h_1(z)\cdot h_2(z)$, which is similar to the histogram based segment similarity measurement in~\cite{ChangLL11}, to force each foreground to be similar with the other foregrounds
and also be different with their backgrounds. The energy generated with
MRF model is proved to be submodular and can be efficiently
solved by GraphCut algorithm. Rubio \emph{et
al.}~\cite{RubioSLP12} evaluated the foreground similarities by
high order graph matching, which is introduced into MRF model to
form the global term. Batra \emph{et al.}~\cite{BatraKPLC10}
firstly proposed an interactive cosegmentation, where an automatic
recommendation system was developed to guide the user to scribble
the uncertain regions for cosegmentation refinement. Several cues
such as uncertainty based cues, scribble based cues and image
level cues are combined to form a recommendation map, where the
regions with larger values are suggested to be labeled by the user.

Apart from MRF segmentation model, Collins \emph{et
al.}~\cite{CollinsXGS12} extended RandomWalk model to solve cosegmentation, which results in a
convex minimization problem with box constraints and a GPU implementation. In~\cite{MengLL12}, active contour based
segmentation is extended for cosegmentation, which consists
of the foreground consistencies across images
and the background consistencies within each image. Due to the linear similarity measurement, the minimization can be resolved by level set
technique.

\subsection{New cosegmentation models}
The second category try to solve cosegmentation problem
using new strategies rather than extending existing single
segmentation models. For example, by treating the common object extraction task as a common region clustering problem, the cosegmentation
problem can be solved by clustering strategy. Joulin \emph{et
al.}~\cite{JoulinBP10} treated the cosegmentation labeling as
training data, and trained a supervised classifier based on the
labeling to see if the given labels are able to result in maximal
separation of the foreground and background classes. The
cosegmentation is then formulated as searching the labels that
lead to the best classification. Spectral method (Normalized Cuts)
is adopted for the bottom-up clustering, and discriminative
clustering is used to share the bottom-up clustering among
images. The two clusterings are finally combined to form the
cosegmentation model, which is solved by convex relaxation and
efficient low-rank optimization. Kim \emph{et al.}~\cite{KimLH12}
solved cosegmentation by clustering strategy, which divides the
images into hierarchical superpixel layers and describes the
relationship of the superpixels using graph. Affinity matrix
considering intra-image edge affinity and inter-image edge
affinity is constructed. The cosegmentation can then be solved by
spectral clustering strategy.

By representing the region similarity relationships as edge
weights of a graph, graph theory has also been used to solve
cosegmentation. In~\cite{VicenteRK11}, by representing each image as a set of object
proposals, a random forest regression based model is learned to
select the object from backgrounds. The relationships between the
foregrounds are formulated by fully connected graph, and the
cosegmentation is achieved by loop belief propagation. Meng
\emph{et al.}~\cite{MengLLN12} constructed directed graph structure
to describe the foreground relationship by only considering the
neighbouring images. The object cosegmentation is then formulated
as s shortest path problem, which can be solved by dynamic
programming.

Some methods try to learn the prior of common objects, which is
then used to segment the common objects. Sun \emph{et
al.}~\cite{SunP13} solved cosegmentation by learning
discriminative part detectors of the object. By forming the
positive and negative training parts from the given images and
other images, the discriminative parts are learned based on the
fact that the part detector of the common objects should more
frequently appear in positive samples than negative samples. The
problem is finally formulated as a latent SVM model learning
problem with group sparsity regularization. Dai \emph{et
al.}~\cite{DaiWZZ13} proposed coskech model by extending the
active basis model to solve cosegmentation problem. In cosketch, a
deformable shape template represented by codebook is generated to
align and extract the common object. The template is introduced
into unsupervised learning process to iteratively sketch the
images based on the shape and segment cues and re-learn the shape
and segment templates with the model parameters. By cosketch,
similar shape objects can be well segmented.

There are also some other strategies. Faktor \emph{et
al.}~\cite{FaktorI13} solved cosegmentation based on the
similarity of composition, where the likelihood of the
co-occurring region is high if it is non-trivial and has good
match with other compositions. The co-occurring regions between
images are firstly detected. Then, consensus scoring by
transferring information among the regions is performed to obtain
cosegments from co-occurring regions. Mukherjee \emph{et
al.}~\cite{MukherjeeSP11} evaluated the foreground similarity by
forcing low entropy of the matrix comprised by the foreground
features, which can handle the scale variation very well.

  %----------------------------------------------------------------------
  \begin{figure*}
    \begin{center}
        \includegraphics[width = 1\columnwidth]{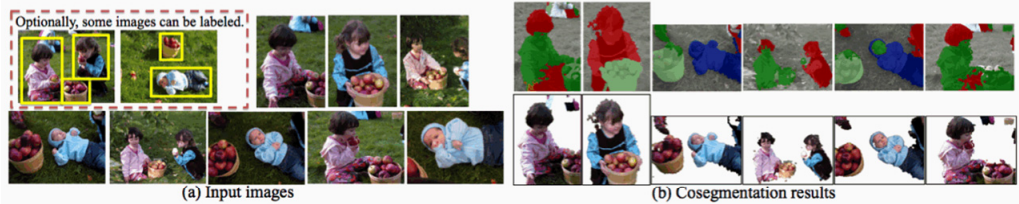}
    \end{center}
    \caption
    {\label{fig:mfc} Given a user's photo stream about certain event, which consists of finite objects, e.g. red-girl, blue-girl, blue-baby and apple basket. Each image contains an unknown subset of them, which we called ``{\emph{Multiple Foreground Cosegmentation}}'' problem. Kim and Xing~\cite{KimX12} proposed the first method to extract irregularly occurred objects from the photo stream (b) by using a few bounding boxes provided by the user in (a).
    }
  \end{figure*}
  %------------------------

\subsection{New cosegmentation problems}
Many applications require the cosegmentation on a large-scale set of
images, which is extremely time consuming. Kim \emph{et al.}~\cite{KimXLK11}
solved the large-scale cosegmentation by temperature maximization
on anisotropic heat diffusion (called CoSand), which starts with
finite sources and performs heat diffusion among images based on
superpixels. The model is submodular with fast solver. Wang
\emph{et al.}~\cite{WangRL13} proposed a semi-supervised learning
based method for large scale image cosegmentation. With very
limited training data, the cosegmentation is formed based on the terms
of inter-image distance to measure the foreground consistencies
among images, the intra-image distance to evaluate the
segmentation smoothness within each image and the balance term to avoid the
same label of all superpixels. The model is converted and
approximated to a convex QP problem and can be solved in
polynomial time using active set. Zhu \emph{et
al.}~\cite{ZhuCZWM13} proposed the first method which uses search
engine to retrieve similar images to the input image to analyze
the object-of-interest information, and then uses the information
to cut-out the object-of-interest. Rubinstein \emph{et
al.}~\cite{RubinsteinJKL13} observed that there are always noise
images (which do not contain the common objects) from web image dataset,
and proposed a cosegmentation model to avoid the noise images. The
main idea is to match the foregrounds using SIFT flow to decide
which images do not contain the common objects.

Applying cosegmentation to improve image classification is an
important application of cosegmentation.
 Chai \emph{et al.}~\cite{ChaiLZ11} proposed a bi-level cosegmentation
method, and used it for image classification. It consists of
bottom level obtained by GrabCut algorithm to initialize the
foregrounds and the top level with a discriminative classification
to propagate the information. Later, a TriCos
model~\cite{ChaiRLGZ12} containing three levels were further proposed, including image level (GrabCut), dataset-level, and category level, which
outperforms the bi-level model.

The sea of personal albums over internet, which depicts
events in short periods, makes one popular source of large image data
that need further analysis. Typical scenario of a personal album is
that it contains multiple objects of different categories in the
image set and each image can contain a subset of them. This is different from the classical cosegmentation models introduced above,
which usually assume each image contains the same common
foreground while having strong variations in backgrounds. The problem of extracting multiple objects from
personal albums is called ``\textit{Multiple Foreground
Cosegmentation}'' (MFC) (see Fig.~\ref{fig:mfc}). Kim and Xing~\cite{KimX12} proposed the
first method to handle MFC problem. Their method starts from
building apperance models (GMM \& SPM) from user-provided bounding
boxes which enclose the object-of-interest, and over-segments the
images into superpixels by~\cite{KimXLK11}. Then they used beam
search to find proposal candidates for each foreground. Finally,
the candidates are seamed into non-overlap regions by using
dynamic programming. Ma \emph{et al.}~\cite{MaL13} formulated the
multiple foreground cosegmentation as semi-supervised learning
(graph transduction learning), and introduced connectivity
constraints to enforce the extraction of connected regions. A
cutting-plane algorithm was designed to efficiently minimize the
model in polynomial time.

Kim and Ma's methods hold an implicit constraint on objects using
low-level cues, and therefore their method might assign labels of
``stuff" (grass, sky) to ``thing'' (people or other objects).
Zhu \emph{et al.}~\cite{Zhu14} proposed a principled CRF framework which
explicitly expresses the object constraints from object detectors
and solves an even more challenging problem: \textit{multiple
foreground recognition and cosegmentation} (MFRC). They proposed an
extended multiple color-line based object detector which can be
on-line trained by using user-provided bounding boxes to detect
objects in unlabeled images. Finally, all the cues from bottom-up
pixels, middle-level contours and high-level object detectors
are integrated in a robust high-order CRF model, which can
enforce the label consistency among pixels, superpixels and object
detections simultaneously, produce higher accuracy in
object regions and achieve state-of-the-art performance for the MFRC
problem. Later, Zhu \emph{et al.}~\cite{ZhuICIP14} further studied another challenging multiple human identification and
cosegmentation problem, and proposed a novel shape cue which uses geodesic filters~\cite{CriminisiSB08} and joint-bilateral
filters to transform the blurry response maps from multiple
color-line object detectors and poselet models to edge-aligned
shape prior. It leads to promising human identification and
co-segmentation performance.

\section{Dataset and Evaluation Metrics}
\label{sec:dataset}
\subsection{Datasets}
To inspire new methods and objectively evaluate their performance
for certain applications, different datasets and evaluation
metrics have been proposed. Initially, the huge labeling cost limits the
size of the datasets~\cite{MartinFTM01, RotherKB04} (typically in
hundreds of images). Recently, with the popularity of crowdsourcing platform
such as Amazon Merchant Turk (AMT) and LabelMe~\cite{RussellTMF08}, the label cost is
shared over the internet users, which makes large datasets with
millions of images and labels possible. Below we summarize the most
influential datasets which are widely used in the existing
segmentation literature ranging from bottom-up image segmentation
to holistic scene understanding:

\subsubsection{Single image segmentation datasets}
\textbf{Berkeley segmentation benchmark dataset}
(BSDS)~\cite{MartinFTM01} is one of the earliest and largest
datasets for contour detection and single image object-agnostic
segmentation with human annotation. The latest BSDS dataset
contains 200 images for training, 100 images for validation and
the rest 200 images for testing. Each image is annotated
by at least 3 subjects. Though the size of the dataset is small,
it still remains one of the most difficult segmentation datasets as
it contains various object classes with great pose variation,
background clutter and other challenges. It has also been used to
evaluate superpixel segmentation methods. Recently,
Li \emph{et al.}~\cite{LiCAZ13} proposed a new benchmark based on
BSDS, which can evaluate semantic segmentation at object or part level.

\textbf{MSRC-interactive segmentation dataset}~\cite{RotherKB04}
includes 50 images with a single binary ground-truth for
evaluating interactive segmentation accuracy.  This dataset also
provides imitated inputs such as labeling-lasso and rectangle with
labels for background, foreground and unknown areas.

\subsubsection{Cosegmentation datasets}
\textbf{MSRC-cosegmentation dataset}~\cite{RotherMBK06} has been
used to evaluate image-pair binary cosegmentation. The dataset
contains 25 image pairs with similar foreground objects but
heterogeneous backgrounds, which matches the assumptions of early
cosegmentation methods~\cite{RotherMBK06,
HochbaumS09,MukherjeeSD09}. Some pairs of the images are picked
such that they contain some camouflage to balance database bias
which forms the baseline cosegmentation dataset.

\textbf{iCoseg dataset}~\cite{BatraKPLC10} is a large binary-class
image cosegmentation dataset for more realistic scenarios. It
contains 38 groups with a total of 643 images. The content of the
images ranges from wild animals, popular landmarks, sports teams to
other groups containing similar foregrounds. Each group contains
images of similar object instances from different poses with some
variations in the background. iCoseg is challenging because the
objects are deformed considerably in terms of viewpoint and
illumination, and in some cases, only a part of the object is
visible. This contrasts significantly with the restrictive scenario of
MSRC-Cosegmentation dataset.

\textbf{FlickrMFC dataset}~\cite{KimX12} is the only dataset for
multiple foreground cosegmentation, which consists of 14 groups of
images with manually labeled ground-truth. Each group includes
10$\sim$20 images which are sampled from a Flikcr photostream. The
image content covers daily scenarios such as children-playing,
fishing, sports, etc. This dataset is perhaps the most
challenging cosegmentation dataset as it contains a number of
repeating subjects that are not necessarily presented in every
image and there are strong occlusions, lighting variations, or
scale or pose changes. Meanwhile, serious background clutters and
variations often make even state-of-the-art object detectors failing on
these realistic scenarios.

\subsubsection{Video segmentation/cosegmentation datasets}
\textbf{SegTrack dataset}~\cite{TsaiBMVC10} is a large
binary-class video segmentation
 dataset with pixel-level annotation for primary objects.
 It contains six videos (bird, bird-fall, girl, monkey-dog, parachute and penguin).
 The dataset contains challenging cases including foreground/background occlusion, large shape deformation and camera motion.

 \textbf{CVC binary-class video cosegmentation dataset}~\cite{RubioSL12} contains 4 synthesis videos which paste the same foreground to different
 backgrounds and 2 videos sampled from the SegTrack. It forms a restrictive
 dataset for early video cosegmentation methods.

 \textbf{MPI multi-class video cosegmentation dataset}~\cite{ChiuF13} was
 proposed to evaluate video cosegmentation approaches in
 more challenging scenarios, which contain multi-class objects. This
 challenging dataset contains 4 different video sets sampled from Youtube
 including 11 videos with around 520 frames with ground truth. Each video
 set has different numbers of object classes appearing in them.
 Moreover, the dataset includes challenging lighting, motion blur and image condition variations.

 \textbf{Cambridge-driving video dataset} (CamVid)~\cite{BrostowSFC2008} is a collection of videos with labels of 32 semantic classes (e.g. building, tree, sidewalk, traffic light, etc), which are captured by a position-fixed CCTV-camera on a driving automobile over 10 mins at 30Hz footage. This dataset contains 4 video sequences, with more than 700 images at resolution of $960\times 720$. Three of them are sampled at the day light condition and the remaining one is sampled at the dark. The number and the heterogeneity of the object classes in each video sequence are diverse.

\subsubsection{Static scene parsing datasets}
\textbf{MSRC 23-class dataset}~\cite{ShottonWRC09} consists of 23 classes and 591
images. Due to the rarity of `horse' and `mountain' classes, these
two classes are often ignored for training and evaluation. The
remaining 21 classes contain diverse objects. The
annotated ground-truth is quite rough.

\textbf{PASCAL VOC dataset}~\cite{EveringhamGWWZ10} provides a
large-scale dataset for evaluating object detection and semantic
segmentation. Starting from the initial 4-class objects in 2005, now
PASCAL dataset includes 20 classes of objects under four major
categories (animal, person, vehicle and indoor). The latest
train/val dataset has 11,530 images and 6,929 segmentations.

\textbf{LabelMe + SUN dataset}: LabelMe~\cite{RussellTMF08} is
initiated by the MIT CSCAIL which provides a dataset of annotated
images. The dataset is still growing. It contains copyright-free
images and is open to public contribution. As of October 31, 2010,
LabelMe has 187,240 images, 62,197 annotated images, and 658,992
labeled objects.  SUN~\cite{XiaoHEOT10} is a subsampled dataset
from LabelMe. It contains 45,676 image (21,182 indoor and 24,494
outdoor), total 515 object categories. One
noteworthy point is that the number of objects in each class is
uneven, which can cause unsatisfactory segmentation accuracy
for rare classes.

\textbf{SIFT-flow dataset}: The popularity of nonparametric scene
parsing requires a large labeled dataset. The SIFT Flow
dataset~\cite{LiuYT11} is composed of
 2,688 images that have been thoroughly labelled by LabelMe users.
 Liu \emph{et al.}~\cite{LiuYT11} have split this dataset into 2,488 training images and 200 test
images and used synonym correction to obtain 33 semantic labels.

\textbf{Stanford background dataset}~\cite{GouldFK09} consists of
around 720 images sampled from the existing datasets such as LabelMe, MSRC
and PASCAL VOC, whose content consists of rural,
urban and harbor scenes. Each image pixel is given two labels:
one for its semantic class (sky, tree, road, grass, water, building,
mountain and foreground) and the one for geometric property (sky, vertical, and
horizontal).

\textbf{NYU dataset}~\cite{Silberman12}: The NYU-depth V2 dataset
is comprised of video sequences from a variety of indoor scenes
recorded by both the RGB and depth cameras of Microsoft
Kinect. It features 1449 densely labeled pairs of aligned RGB and
depth images, 464 new scenes taken from 3 cities, and 407,024 new
unlabeled frames. Each object is labeled with a class and an
instance number (cup1, cup2, cup3, etc).

\textbf{Microsoft COCO}~\cite{LinMBHPRDZ14} is a recent dataset
for holistic scene understanding, which provides 328K images
with 91 object classes. One substantial difference with other
large datasets, such as PASCAL VOC and SUN datasets, is that
Microsoft COCO contains more labelled instances in million units. The
authors argue that it can facilitate training object detectors with
better localization, and learning contextual information.

\subsection{Evaluation metrics}
%% Begin evaluation metric parts
As segmentation is an ill-defined problem, how to evaluate an
algorithm's goodness remains an open question. In the past, the
evaluation were mainly conducted through subjective human
inspections or by evaluating the performance of subsequent vision
system which uses image segmentation. To objectively evaluate a method, it is
desirable to associate the segmentation with perceptual
grouping. Current trend is to develop a
benchmark~\cite{MartinFTM01} which consists of human-labeled
segmentation and then compares the algorithm's output with the
human-labeled results using some metrics to measure the
segmentation quality. Various evaluation metrics have been
proposed:

\begin{itemize}
    \item \textbf{Boundary Matching}: This method works by matching the
    algorithm-generated boundaries with human-labeled boundaries, and then
    computing some metric to evaluate the matching quality. Precision and recall
    framework proposed by Martin \emph{et al.} \cite{Martin:CSD-03-1268} is among
    the widely used evaluation metrics, where the \emph{Precision} measures
    the proportion of how many machine-generated boundaries can be found
    in human-labeled boundaries and is sensitive to over-segmentation, while
    the \emph{Recall} measures the proportion of how many human-labelled boundaries can be found in machine-generated boundaries and
    is sensitive to under-segmentation. In general, this method is sensitive
    to the granularity of human labeling.
    \item \textbf{Region Covering }: This method \cite{Martin:CSD-03-1268} operates by checking
    the overlap between the machine-generated regions and human-labelled
    regions. Let $S_{seg}$ and $S_{hum}$ denote the machine segmentation
    and the human segmentation, respectively. Denote the corresponding segment regions for pixel
    $p_i$ from the pixel set $P=\{p_1, p_2, ..., p_n\}$ as $C(S_{seg}, p_i)$
    and $C(S_{hum}, p_i)$. The relative region covering error at $p_i$ is
    \begin{equation}\label{region overlap}
    O(S_{seg}, S_{label}, p_i)= \frac{|C(S_{seg}, p_i)\setminus C(S_{hum}, p_i)|}{C(S_{seg},
    p_i)}.
    \end{equation}
    where $\setminus$ is the set differencing operator.
    
    The globe region covering error is defined as:
    \begin{equation}\label{total_overlap}
    \begin{split}
    & GCE(S_{seg}, S_{label})= \\
    & \frac{1}{n}min\{\sum_{i}O(S_{seg}, S_{label}, p_i),O(S_{label}, S_{seg},
    p_i)\}.
    \end{split}
    \end{equation}

    However, when each pixel is a segment or the whole image is a segment,
    the $GCE$ becomes zero which is undesirable. To alleviate these problems,
    the authors proposed to replace $min$ operation by using $max$ operation, but
    such change will not encourage segmentation at finer detail. 
    
    Another commonly used region based criterion
    is the Intersection-over-Union, by checking the overlap between the $S_{seg}$ and $S_{hum}$:
    \begin{equation}\label{IoU}
	    IoU(S_{seg}, S_{hum}) = \frac{S_{Seg} \cap S_{hum}}{S_{Seg} \cup S_{hum}}
    \end{equation}

    \item \textbf{Variation of Information} (VI): This metric~\cite{Meila03}
    measures the distance between two segmentations $S_{seg}$ and $S_{hum}$ using
    average conditional entropy. Assume that $S_{seg}$ and $S_{hum}$
    have
    clusters $C_{seg}=\{C_{seg}^1, C_{seg}^2,..., C_{seg}^K\}$ and $C_{hum}=\{C_{hum}^1, C_{hum}^2,..., C_{hum}^{K'}\}$,
    respectively. The variation of information is defined as:
    \begin{equation}\label{variation_information}
    VI(S_{seg}, S_{hum})= H(S_{seg}) + H(S_{hum})-2I(S_{seg}, S_{hum})
    \end{equation}
    where $H(S_{seg})$ is the entropy associated with clustering $S_{seg}$:
    \begin{equation}\label{entropy}
    H(S_{seg})=-\sum_{k=1}^K P(k)log P(k)
    \end{equation}
    with $P(k)= \frac{n_k}{n}$, where $n_k$ is the number of elements in cluster $k$ and $n$ is the
    total number of elements in $S_{seg}$. $I(S_{seg}, S_{hum})$ is the mutual
    information between $S_{seg}$ and $S_{hum}$:
    \begin{equation}\label{mutual information}
    I(S_{seg}, S_{hum})=\sum_{k=1}^K\sum_{k'=1}^{K'}P(k,k')log\frac{P(k,k')}{P(k)P(k')}
    \end{equation}
    with $P(k, k')=\frac{|C_{seg}^k \bigcap C_{hum}^k|}{n}$.

    Although $VI$ posses some interesting property, its perceptual
meaning and potential in evaluating more than one ground-truth are
unknown~\cite{ArbelaezMFM09}.
    \item \textbf{Probabilistic Random Index} (PRI): PRI was introduced to measure
    the compatibility of assignments between pairs of elements
    in $S_{seg}$ and $S_{hum}$. It
    has been defined to deal with multiple ground-truth segmentations~\cite{UnnikrishnanPH07}:
    \begin{equation}\label{PRI}
    PRI(S_{seg}, S_{hum}^k)=\frac{1}{\frac{N}{2}}\sum_{i<j}[c_{ij}p_{ij}+(1-c_{ij})(1-p_{ij})]
    \end{equation}
    where $S_{hum}^k$ is the $k$-th human-labeled ground-truth, $c_{ij}$ is the event that pixels $i$ and $j$ have the same label and $p_{ij}$ is
    the corresponding probability. As reported in~\cite{ArbelaezMFM11},
    the PRI has a small dynamic range and the values across images and algorithms are often similar
    which makes the differentiation difficult.
\end{itemize}

\section{Discussions and Future Directions}
\label{sec:discuss}

\subsection{Design flavors}
When designing a new segmentation algorithm, it is often difficult
to make choices among various design flavors, e.g. to use
superpixel or not, to use more images or not. It all depends on
the applications. Our literature review has cast some lights on
the pros and cons of some commonly used design flavors, which is
worth thinking twice before going ahead for a specific setup.

\textbf{Patch vs. region vs. object proposal}: Careful readers
might notice that there has been a significant trend in migrating
from patch based analysis to region based (or superpixel based)
analysis. The continuous performance improvement in terms of
boundary recall and execution time makes superpixel a fast
preprocessing technique. The advantages of using superpixel lie in
not only the time reduction in training and inference but also
more complex and discriminative features that can be exploited. On
the other hand, superpixel itself is not perfect, which could
introduce new structure errors. For users who care more about
visual results on the segment boundary, pixel-based or hybrid
approach of combining pixel and superpixel should be considered as
better options. The structure errors of using superpixel can also
be alleviated by using different methods and parameters to produce
multiple over-segmentation or using fast edge-aware filtering to
refine the boundary. For users more caring about localization
accuracy, the region based way is more preferred due to the
various advantages introduced while the boundary loss can be
neglected. Another uprising trend that is worth mentioning is the application
of the object region proposals~\cite{VicenteRK11, MengLLN12}. Due to
the larger support provided by object-like regions than
oversegmentation or pixels, more complex classifiers and
region-level features can be extracted. However, the recall rate
of the object proposal is still not satisfactory (around $60\%$);
therefore more careful designs need to be made when accuracy is a
major concern.

\textbf{Intrinsic cues vs. extrinsic cues}: Although intrinsic
cues (the features and prior knowledge for a single image) still
play dominant roles in existing CV applications, extrinsic cues
which come from multiple images (such as multiview images,
video sequence, and a super large image dataset of similar
products) are attracting more and more attentions.  An intuitive
answer why extrinsic cues convey more semantics can be interpreted
in terms of statistic. When there are many signals available, the
signals which repeatedly appear form patterns of interest, while
those errors are averaged. Therefore, if there are multiple images containing redundant but
diverse information, incorporating extrinsic cues should bring
some improvements. When taking extrinsic cues, the source of
information needs to be considered in the algorithm design. More robust constraints such as the cues from multiple-view
geometric or spatial-temporal relationships should be exploited first. When working with external
information such as a large dataset which contains heterogeneous
data, a mechanism that can handle noisy information should be
developed.

\textbf{Hand-crafted features vs. learned features}: Hand-crafted
features, such as intensity, color, SIFT and Bag-of-Word, etc,
have played important roles in computer vision. These simple and
training-free features have been applied to many applications, and
their effectiveness has been widely examined.  However, the
generalization capability of these hand-crafted features from one
task to another task depends on the heuristic feature design and
combination, which can compromise the performance. The development of
low-level features has become more and more challenging.  On the
other hand, learned features from labelled database
have recently been demonstrated advantages in some applications, such as
scene understanding and object detection.  The effectiveness of
the learned features comes from the context information captured
from longer spatial arrangement and higher order co-occurrence.
With labelled data, some structured noise is eliminated which
helps highlight the salient structures. However, learned features
can only detect patterns for certain tasks. If migrating to other
tasks, it needs newly labelled data, which is time consuming and
expensive to obtain.  Therefore, it is suggested to choose
features from the handcraft features first. If it happens
to have labelled data, then using learned features usually
boost up the performance.

\subsection{Promising future directions}
Based on the discussed literature and the recent development in
segmentation community, here we suggest some future directions
which is worth for exploring:

\textbf{Beyond single image}: Although single image segmentation
has played a dominant role in the past decades, the recent trend
is to use the internal structures of multiple images to facilitate
segmentation. When human perceives a scene, the motion cues and
depth cues provide extra information, which should not be
neglected. For example, when the image data is from a video, the
3D geometry estimated from the structure-from-motion techniques
can be used to help image understanding~\cite{SturgessALT09,
ZhangWY10}. The depth information captured by commodity depth
cameras such as Kinect also benefits the indoor scene
understanding~\cite{SilbermanF11,SilbermanHKF12}. Beyond 3D
information, the useful information from other 2D images in a
large database, either organized (like ImageNet) or un-organized
(like product or animal category), or multiple heterogeneous
datasets of the same category, can also be
exploited~\cite{KuttelGF12,Fanman2014}.

\textbf{Toward multiple instances}: Most segmentation methods
still aim to produce a most likely solution, e.g. all pixels of
the same category are assigned the same label. On the other hand,
people are also interested in knowing the information of ``how
many'' of the same category objects are there, which is a
limitation of the existing works. Recently, there are some works
making efforts towards this direction by combining
state-of-the-art detectors with the CRF
modelling~\cite{LadickySART10, HeG14,TigheNL14} which have
achieved some progress. In addition, the recently developed
dataset, Microsoft COCO~\cite{LinMBHPRDZ14}, which contains many
images with labelled instances, can be expected to boost the
development in this direction.

\textbf{Become more holistic}: Most existing works consider the
image labelling task alone~\cite{HeZC04, ShottonWRC09}. On the
other hand, studying a related task can improve the existing scene
analysis problem. For example, by combining the object detection
with image classification, Yao \emph{et al.}~\cite{YaoFU12}
proposed a more robust and accurate system than the ones which
perform single task analysis. Such holistic understanding trend
can also be seen from other works, by combining
context~\cite{RabinovichVGWB07, DivvalaHHEH09},
geometry~\cite{GouldFK09, GuptaSEH11, HoiemEH07, VineetWT12},
attributes~\cite{FarhadiEH10, KumarBBN11, LampertNH09, TigheL11,
ZhengCWSVRT14} or even language~\cite{GuptaD08,
KulkarniPODLCBB13}. It is therefore expected that a more holistic
system should lead to better performance than those performing
monotone analysis, though at an increasing inference cost.

\textbf{Go from shallow to deep}: Feature has been played an
important role in many vision applications, e.g. stereo matching,
detection and segmentation. Good features, which are highly
discriminative to differentiate `object' from `stuff', can help
object segmentation significantly. However, most features today
are hand-crafted designed for specific tasks. When applied to
other tasks, the hand-crafted features might not generalize well.
Recently, learned features using multiple-layer neural
network~\cite{KrizhevskySH12} have been applied in many vision
tasks, including object classification~\cite{ErhanSTA13}, face
detection~\cite{ToshevS13}, pose estimation~\cite{SzegedyTE13},
object detection~\cite{GirshickDDM14} and semantic
segmentation~\cite{Farabet2013}, and achieved state-of-the-art
performance. Therefore, it is interesting to explore whether such
learned features can benefit the aforementioned more complex
tasks.

\section{Conclusion}
\label{sec:summary} In this paper, we have conducted a thorough
review of recent development of image segmentation methods,
including classic bottom-up methods, interactive methods, object
region proposals, semantic parsing and image cosegmentation. We
have discussed the pros and cons of different methods, and
suggested some design choices and future directions which are
worthwhile for exploration.

Segmentation as a community has achieved substantial progress in
the past decades. Although some technologies such as interactive
segmentation have been commercialized in recent Adobe and
Microsoft products, there is still a long way to make segmentation
a general and reliable tool to be widely applied to practical
applications. This is more related to the existing methods'
limitations in terms of robustness and efficiency. For example,
one can always observe that given images under various
degradations (strong illumination change, noise corruption or rain
situation), the performance of the existing methods could drop
significantly~\cite{Freeman2011}. With the rapid improvement in
computing hardware, more effective and robust methods will be
developed, where the breakthrough is likely to come from the
collaborations with other engineering and science communities,
such as physics, neurology and mathematics.
\vspace{10pt}

%% can use a bibliography generated by BibTeX as a .bbl file
%% BibTeX documentation can be easily obtained at:
%% http://www.ctan.org/tex-archive/biblio/bibtex/contrib/doc/
%% The IEEEtran BibTeX style support page is at:
%% http://www.michaelshell.org/tex/ieeetran/bibtex/
%\bibliographystyle{IEEEtran}
%\bibliography{reference}

\bibliographystyle{elsarticle-num}
\bibliography{reference}

\begin{thebibliography}{100}
\expandafter\ifx\csname url\endcsname\relax
  \def\url#1{\texttt{#1}}\fi
\expandafter\ifx\csname urlprefix\endcsname\relax\def\urlprefix{URL }\fi
\expandafter\ifx\csname href\endcsname\relax
  \def\href#1#2{#2} \def\path#1{#1}\fi

\bibitem{CarsonBGM02}
C.~Carson, S.~Belongie, H.~Greenspan, J.~Malik, Blobworld: Image segmentation
  using expectation-maximization and its application to image querying, IEEE
  Trans. Pattern Anal. Mach. Intell. 24~(8) (2002) 1026--1038.

\bibitem{AchantaSSLFS12}
R.~Achanta, A.~Shaji, K.~Smith, A.~Lucchi, P.~Fua, S.~S{\"u}sstrunk, Slic
  superpixels compared to state-of-the-art superpixel methods, IEEE Trans.
  Pattern Anal. Mach. Intell. 34~(11).

\bibitem{ComaniciuM02}
D.~Comaniciu, P.~Meer, Mean shift: {A} robust approach toward feature space
  analysis, {IEEE} Trans. Pattern Anal. Mach. Intell. 24~(5) (2002) 603--619.

\bibitem{LuYMD13}
J.~Lu, H.~Yang, D.~Min, M.~N. Do, Patch match filter: Efficient edge-aware
  filtering meets randomized search for fast correspondence field estimation,
  in: CVPR, 2013.

\bibitem{KohliKT09}
P.~Kohli, M.~P. Kumar, P.~H.~S. Torr, P {\&} beyond: Move making algorithms for
  solving higher order functions, IEEE Trans. Pattern Anal. Mach. Intell.
  31~(9).

\bibitem{Mori05}
G.~Mori, Guiding model search using segmentation, in: ICCV, 2005.

\bibitem{GuLAM09}
C.~Gu, J.~J. Lim, P.~Arbelaez, J.~Malik, Recognition using regions, in: CVPR,
  2009.

\bibitem{CarreiraS12}
J.~Carreira, C.~Sminchisescu, Cpmc: Automatic object segmentation using
  constrained parametric min-cuts, IEEE Trans. Pattern Anal. Mach. Intell.
  34~(7).

\bibitem{EndresH14}
I.~Endres, D.~Hoiem, Category-independent object proposals with diverse
  ranking, IEEE Trans. Pattern Anal. Mach. Intell. 36~(2).

\bibitem{ShiM00}
J.~Shi, J.~Malik, Normalized cuts and image segmentation, IEEE Trans. Pattern
  Anal. Mach. Intell. 22~(8) (2000) 888--905.

\bibitem{MartinFM04}
D.~R. Martin, C.~Fowlkes, J.~Malik, Learning to detect natural image boundaries
  using local brightness, color, and texture cues, IEEE Trans. Pattern Anal.
  Mach. Intell.

\bibitem{Hoffman1997}
D.~D. Hoffman, M.~Singh, Salience of visual parts, Cognition 63~(1) (1997) 29
  -- 78.
\newblock \href {http://dx.doi.org/10.1016/S0010-0277(96)00791-3}
  {\path{doi:10.1016/S0010-0277(96)00791-3}}.

\bibitem{Ohlander78}
R.~Ohlander, K.~Price, D.~R. Reddy, Picture segmentation using a recursive
  region splitting method, Computer Graphics and Image Processing 8 (1978)
  313--333.

\bibitem{Brice1970}
C.~R. Brice, C.~L. Fennema, Scene analysis using regions., Artif. Intell. 1~(3)
  (1970) 205--226.

\bibitem{FelzenszwalbH04}
P.~F. Felzenszwalb, D.~P. Huttenlocher, Efficient graph-based image
  segmentation, International Journal of Computer Vision 59~(2) (2004)
  167--181.

\bibitem{VincentS91}
L.~Vincent, P.~Soille, Watersheds in digital spaces: An efficient algorithm
  based on immersion simulations, IEEE Trans. Pattern Anal. Mach. Intell.
  13~(6) (1991) 583--598.

\bibitem{Beare06}
R.~Beare, A locally constrained watershed transform, IEEE Trans. Pattern Anal.
  Mach. Intell. 28~(7) (2006) 1063--1074.

\bibitem{Chan2001}
T.~F. Chan, L.~A. Vese, Active contours without edges, IEEE Transactions on
  Image Processing 10~(2) (2001) 266--277.

\bibitem{Osher88}
S.~Osher, J.~A. Sethian, Fronts propagating with curvature dependent speed:
  Algorithms based on hamilton-jacobi formulations, Journal of Computational
  Physics 79~(1) (1988) 12--49.

\bibitem{RotherKB04}
C.~Rother, V.~Kolmogorov, A.~Blake, Grabcut: interactive foreground extraction
  using iterated graph cuts, ACM Trans. Graph. 23~(3).

\bibitem{YangCZL10}
W.~Yang, J.~Cai, J.~Zheng, J.~Luo, User-friendly interactive image segmentation
  through unified combinatorial user inputs, IEEE Transactions on Image
  Processing 19~(9) (2010) 2470--2479.

\bibitem{AnhCZZ12}
T.~N.~A. Nguyen, J.~Cai, J.~Zhang, J.~Zheng, Robust interactive image
  segmentation using convex active contours, {IEEE} Transactions on Image
  Processing 21~(8) (2012) 3734--3743.

\bibitem{ShottonWRC09}
J.~Shotton, J.~M. Winn, C.~Rother, A.~Criminisi, Textonboost for image
  understanding: Multi-class object recognition and segmentation by jointly
  modeling texture, layout, and context, International Journal of Computer
  Vision 81~(1).

\bibitem{FelzenszwalbGM10}
P.~F. Felzenszwalb, R.~B. Girshick, D.~A. McAllester, Cascade object detection
  with deformable part models, in: CVPR, 2010.

\bibitem{HoiemEH08}
D.~Hoiem, A.~A. Efros, M.~Hebert, Putting objects in perspective.,
  International Journal of Computer Vision 80~(1) (2008) 3--15.

\bibitem{KohliLT09}
P.~Kohli, L.~Ladicky, P.~H.~S. Torr, Robust higher order potentials for
  enforcing label consistency, International Journal of Computer Vision 82~(3).

\bibitem{GouldFK09}
S.~Gould, R.~Fulton, D.~Koller, Decomposing a scene into geometric and
  semantically consistent regions, in: ICCV, 2009.

\bibitem{LiuYT11a}
C.~Liu, J.~Yuen, A.~Torralba, Nonparametric scene parsing via label transfer,
  IEEE Trans. Pattern Anal. Mach. Intell. 33~(12).

\bibitem{TigheL13}
J.~Tighe, S.~Lazebnik, Finding things: Image parsing with regions and
  per-exemplar detectors, in: CVPR, 2013.

\bibitem{KrahenbuhlK11}
P.~Kr{\"a}henb{\"u}hl, V.~Koltun, Efficient inference in fully connected crfs
  with gaussian edge potentials, in: NIPS, 2011, pp. 109--117.

\bibitem{LadickyRKT10}
L.~Ladicky, C.~Russell, P.~Kohli, P.~H.~S. Torr, Graph cut based inference with
  co-occurrence statistics, in: ECCV, 2010.

\bibitem{DengDSLL009}
J.~Deng, W.~Dong, R.~Socher, L.-J. Li, K.~Li, F.-F. Li, Imagenet: A large-scale
  hierarchical image database, in: CVPR, 2009.

\bibitem{KimX12}
G.~Kim, E.~P. Xing, On multiple foreground cosegmentation, in: CVPR, 2012, pp.
  837--844.

\bibitem{KimXLK11}
G.~Kim, E.~P. Xing, F.-F. Li, T.~Kanade, Distributed cosegmentation via
  submodular optimization on anisotropic diffusion, in: ICCV, 2011, pp.
  169--176.

\bibitem{JoulinBP12}
A.~Joulin, F.~Bach, J.~Ponce, Multi-class cosegmentation, in: CVPR, 2012.

\bibitem{RubioSLP12}
J.~C. Rubio, J.~Serrat, A.~M. L{\'o}pez, N.~Paragios, Unsupervised
  co-segmentation through region matching, in: CVPR, 2012.

\bibitem{JoulinBP10}
A.~Joulin, F.~R. Bach, J.~Ponce, Discriminative clustering for image
  co-segmentation, in: CVPR, 2010, pp. 1943--1950.

\bibitem{ChangLL11}
K.-Y. Chang, T.-L. Liu, S.-H. Lai, From co-saliency to co-segmentation: An
  efficient and fully unsupervised energy minimization model, in: CVPR, 2011,
  pp. 2129--2136.

\bibitem{MukherjeeSP11}
L.~Mukherjee, V.~Singh, J.~Peng, Scale invariant cosegmentation for image
  groups, in: CVPR, 2011.

\bibitem{MukherjeeSD09}
L.~Mukherjee, V.~Singh, C.~R. Dyer, Half-integrality based algorithms for
  cosegmentation of images, in: CVPR, 2009.

\bibitem{Szeliski11}
R.~Szeliski, Computer Vision - Algorithms and Applications, Texts in Computer
  Science, Springer, 2011.

\bibitem{RaoMYSM09}
S.~Rao, H.~Mobahi, A.~Y. Yang, S.~Sastry, Y.~Ma, Natural image segmentation
  with adaptive texture and boundary encoding, in: ACCV, 2009.

\bibitem{ShiM97}
J.~Shi, J.~Malik, Normalized cuts and image segmentation, in: CVPR, 1997.

\bibitem{MalikBSL99}
J.~Malik, S.~Belongie, J.~Shi, T.~K. Leung, Textons, contours and regions: Cue
  integration in image segmentation, in: ICCV, 1999.

\bibitem{ChengLWHY11}
B.~Cheng, G.~Liu, J.~Wang, Z.~Huang, S.~Yan, Multi-task low-rank affinity
  pursuit for image segmentation, in: ICCV, 2011.

\bibitem{NgJW01}
A.~Y. Ng, M.~I. Jordan, Y.~Weiss, On spectral clustering: Analysis and an
  algorithm, in: NIPS, 2001, pp. 849--856.

\bibitem{CourBS05}
T.~Cour, F.~B{\'e}n{\'e}zit, J.~Shi, Spectral segmentation with multiscale
  graph decomposition, in: CVPR, 2005, pp. 1124--1131.

\bibitem{ArbelaezMFM09}
P.~Arbelaez, M.~Maire, C.~C. Fowlkes, J.~Malik, From contours to regions: An
  empirical evaluation, in: CVPR, 2009.

\bibitem{JainTLB04}
A.~K. Jain, A.~P. Topchy, M.~H.~C. Law, J.~M. Buhmann, Landscape of clustering
  algorithms, in: ICPR (1), 2004, pp. 260--263.

\bibitem{ArbelaezMFM11}
P.~Arbelaez, M.~Maire, C.~Fowlkes, J.~Malik, Contour detection and hierarchical
  image segmentation, IEEE Trans. Pattern Anal. Mach. Intell. 33~(5) (2011)
  898--916.

\bibitem{BressonEVTO07}
X.~Bresson, S.~Esedoglu, P.~Vandergheynst, J.-P. Thiran, S.~Osher, Fast global
  minimization of the active contour/snake model, Journal of Mathematical
  Imaging and Vision 28~(2) (2007) 151--167.

\bibitem{PockCCB09}
T.~Pock, A.~Chambolle, D.~Cremers, H.~Bischof, A convex relaxation approach for
  computing minimal partitions, in: CVPR, 2009, pp. 810--817.

\bibitem{YuanBTB10}
J.~Yuan, E.~Bae, X.-C. Tai, Y.~Boykov, A continuous max-flow approach to potts
  model, in: ECCV, 2010, pp. 379--392.

\bibitem{AmbrosioTOR90}
L.~Ambrosio, V.~Tortorelli, Approximation of functionals depending on jumps by
  elliptic functionals via {$\Gamma$}-convergence, Communications on Pure and
  Applied Mathematics XLIII (1990) 999--1036.

\bibitem{Braides97}
A.~Braides, G.~Dal~Maso, Non-local approximation of the mumford-shah
  functional, Calculus of Variations and Partial Differential Equations 5~(4)
  (1997) 293--322.
\newblock \href {http://dx.doi.org/10.1007/s005260050068}
  {\path{doi:10.1007/s005260050068}}.

\bibitem{Braides98a}
A.~Braides, Approximation of Free-Discontinuity Problems, Vol. 1694 of Lecture
  Notes in Mathematics, Springer, 1998.

\bibitem{Chambolle95}
A.~Chambolle, Image segmentation by variational methods: Mumford and shah
  functional and the discrete approximations, SIAM J. Appl. Math. 55~(3) (1995)
  827--863.
\newblock \href {http://dx.doi.org/10.1137/S0036139993257132}
  {\path{doi:10.1137/S0036139993257132}}.

\bibitem{Vese2002}
L.~A. Vese, T.~F. Chan, A multiphase level set framework for image segmentation
  using the mumford and shah model, International Journal of Computer Vision
  50~(3) (2002) 271--293.

\bibitem{Kass1988}
M.~Kass, A.~Witkin, D.~Terzopoulos, Snakes: Active contour models,
  International Journal of Computer Vision 1~(4) (1988) 321--331.
\newblock \href {http://dx.doi.org/10.1007/BF00133570}
  {\path{doi:10.1007/BF00133570}}.

\bibitem{ZhuZCM13}
H.~Zhu, J.~Zheng, J.~Cai, N.~Magnenat{-}Thalmann, Object-level image
  segmentation using low level cues, {IEEE} Transactions on Image Processing
  22~(10) (2013) 4019--4027.

\bibitem{RenM03}
X.~Ren, J.~Malik, Learning a classification model for segmentation, in: ICCV,
  2003, pp. 10--17.

\bibitem{LevinshteinSKFDS09}
A.~Levinshtein, A.~Stere, K.~N. Kutulakos, D.~J. Fleet, S.~J. Dickinson,
  K.~Siddiqi, Turbopixels: Fast superpixels using geometric flows, {IEEE}
  Trans. Pattern Anal. Mach. Intell. 31~(12) (2009) 2290--2297.

\bibitem{WangZGWZ13}
P.~Wang, G.~Zeng, R.~Gan, J.~Wang, H.~Zha, Structure-sensitive superpixels via
  geodesic distance, International Journal of Computer Vision 103~(1) (2013)
  1--21.

\bibitem{VekslerBM10}
O.~Veksler, Y.~Boykov, P.~Mehrani, Superpixels and supervoxels in an energy
  optimization framework, in: Computer Vision - {ECCV} 2010 - 11th European
  Conference on Computer Vision, Heraklion, Crete, Greece, September 5-11,
  2010, Proceedings, Part {V}, 2010, pp. 211--224.

\bibitem{ZhangHMB11}
Y.~Zhang, R.~I. Hartley, J.~Mashford, S.~Burn, Superpixels via pseudo-boolean
  optimization, in: ICCV, 2011, pp. 1387--1394.

\bibitem{Liu2014}
M.~Liu, O.~Tuzel, S.~Ramalingam, R.~Chellappa, Entropy-rate clustering: Cluster
  analysis via maximizing a submodular function subject to a matroid
  constraint, {IEEE} Trans. Pattern Anal. Mach. Intell. 36~(1) (2014) 99--112.

\bibitem{BerghBRCG12}
M.~V. den Bergh, X.~Boix, G.~Roig, B.~de~Capitani, L.~J.~V. Gool, {SEEDS:}
  superpixels extracted via energy-driven sampling, in: ECCV, 2012, pp. 13--26.

\bibitem{McGuinnessO10}
K.~McGuinness, N.~E. O'Connor, A comparative evaluation of interactive
  segmentation algorithms, Pattern Recognition 43~(2) (2010) 434--444.

\bibitem{Yi2012}
F.~Yi, I.~Moon, Image segmentation: A survey of graph-cut methods, in: Systems
  and Informatics (ICSAI), 2012 International Conference on, 2012.

\bibitem{He2013}
J.~He, C.-S. Kim, C.-C.~J. Kuo, Interactive Segmentation Techniques: Algorithms
  and Performance Evaluation, Springer Publishing Company, Incorporated, 2013.

\bibitem{KassWT88}
M.~Kass, A.~P. Witkin, D.~Terzopoulos, Snakes: Active contour models,
  International Journal of Computer Vision 1~(4) (1988) 321--331.

\bibitem{Mortensen1992}
E.~Mortensen, B.~Morse, W.~Barrett, J.~Udupa, Adaptive boundary detection using
  `live-wire' two-dimensional dynamic programming, in: Computers in Cardiology
  1992, Proceedings of, 1992, pp. 635--638.

\bibitem{BoykovJ01}
Y.~Boykov, M.-P. Jolly, Interactive graph cuts for optimal boundary and region
  segmentation of objects in n-d images, in: ICCV, 2001, pp. 105--112.

\bibitem{MortensenB95}
E.~N. Mortensen, W.~A. Barrett, Intelligent scissors for image composition, in:
  {SIGGRAPH}, 1995, pp. 191--198.

\bibitem{LiuY12}
Y.~Liu, Y.~Yu, Interactive image segmentation based on level sets of
  probabilities, {IEEE} Trans. Vis. Comput. Graph. 18~(2) (2012) 202--213.

\bibitem{Grady06}
L.~Grady, Random walks for image segmentation, IEEE Trans. Pattern Anal. Mach.
  Intell. 28~(11) (2006) 1768--1783.

\bibitem{LiSTS04}
Y.~Li, J.~Sun, C.~Tang, H.~Shum, Lazy snapping, {ACM} Trans. Graph. 23~(3)
  (2004) 303--308.

\bibitem{KohliOJ13}
P.~Kohli, A.~Osokin, S.~Jegelka, A principled deep random field model for image
  segmentation, in: CVPR, 2013, pp. 1971--1978.

\bibitem{PriceMC10a}
B.~L. Price, B.~S. Morse, S.~Cohen, Geodesic graph cut for interactive image
  segmentation, in: CVPR, 2010, pp. 3161--3168.

\bibitem{BaiS07}
X.~Bai, G.~Sapiro, A geodesic framework for fast interactive image and video
  segmentation and matting, in: ICCV, 2007, pp. 1--8.

\bibitem{Veksler08}
O.~Veksler, Star shape prior for graph-cut image segmentation, in: Computer
  Vision - {ECCV} 2008, 10th European Conference on Computer Vision, Marseille,
  France, October 12-18, 2008, Proceedings, Part {III}, 2008, pp. 454--467.

\bibitem{GulshanRCBZ10}
V.~Gulshan, C.~Rother, A.~Criminisi, A.~Blake, A.~Zisserman, Geodesic star
  convexity for interactive image segmentation, in: The Twenty-Third {IEEE}
  Conference on Computer Vision and Pattern Recognition, {CVPR} 2010, San
  Francisco, CA, USA, 13-18 June 2010, 2010, pp. 3129--3136.

\bibitem{CriminisiSB08}
A.~Criminisi, T.~Sharp, A.~Blake, Geos: Geodesic image segmentation, in:
  Computer Vision - {ECCV} 2008, 10th European Conference on Computer Vision,
  Marseille, France, October 12-18, 2008, Proceedings, Part {I}, 2008, pp.
  99--112.

\bibitem{VicenteKR08}
S.~Vicente, V.~Kolmogorov, C.~Rother, Graph cut based image segmentation with
  connectivity priors, in: CVPR, 2008.

\bibitem{HosniRBRG13}
A.~Hosni, C.~Rhemann, M.~Bleyer, C.~Rother, M.~Gelautz, Fast cost-volume
  filtering for visual correspondence and beyond, {IEEE} Trans. Pattern Anal.
  Mach. Intell. 35~(2) (2013) 504--511.

\bibitem{DurandD02}
F.~Durand, J.~Dorsey, Fast bilateral filtering for the display of
  high-dynamic-range images, {ACM} Trans. Graph. 21~(3) (2002) 257--266.

\bibitem{He0T13}
K.~He, J.~Sun, X.~Tang, Guided image filtering, {IEEE} Trans. Pattern Anal.
  Mach. Intell. 35~(6) (2013) 1397--1409.

\bibitem{LuSMLD12}
J.~Lu, K.~Shi, D.~Min, L.~Lin, M.~N. Do, Cross-based local multipoint
  filtering, in: CVPR, 2012, pp. 430--437.

\bibitem{Adams1994}
R.~Adams, L.~Bischof, Seeded region growing, IEEE Trans. Pattern Anal. Mach.
  Intell. 16~(6) (1994) 641--647.

\bibitem{NingZZW10}
J.~Ning, L.~Zhang, D.~Zhang, C.~Wu, Interactive image segmentation by maximal
  similarity based region merging, Pattern Recognition 43~(2) (2010) 445--456.

\bibitem{BourdevMBM10}
L.~D. Bourdev, S.~Maji, T.~Brox, J.~Malik, Detecting people using mutually
  consistent poselet activations, in: ECCV, 2010.

\bibitem{LarlusJ08}
D.~Larlus, F.~Jurie, Combining appearance models and markov random fields for
  category level object segmentation, in: CVPR, 2008.

\bibitem{AlexeDF12}
B.~Alexe, T.~Deselaers, V.~Ferrari, Measuring the objectness of image windows,
  {IEEE} Trans. Pattern Anal. Mach. Intell. 34~(11) (2012) 2189--2202.

\bibitem{Hosang14}
J.~Hosang, R.~Benenson, B.~Schiele, How good are detection proposals, really?,
  in: British Machine Vision Conference (BMVC), British Machine Vision
  Association, British Machine Vision Association, 2014.

\bibitem{BorjiCJL14}
A.~Borji, M.~Cheng, H.~Jiang, J.~Li, Salient object detection: {A} survey, CoRR
  abs/1411.5878.

\bibitem{HoiemEH11}
D.~Hoiem, A.~A. Efros, M.~Hebert, Recovering occlusion boundaries from an
  image, International Journal of Computer Vision 91~(3) (2011) 328--346.

\bibitem{HumayunLR14}
A.~Humayun, F.~Li, J.~M. Rehg, {RIGOR:} reusing inference in graph cuts for
  generating object regions, in: CVPR, 2014, pp. 336--343.

\bibitem{KohliRBT08}
P.~Kohli, J.~Rihan, M.~Bray, P.~H.~S. Torr, Simultaneous segmentation and pose
  estimation of humans using dynamic graph cuts, International Journal of
  Computer Vision 79~(3) (2008) 285--298.

\bibitem{KrahenbuhlK14}
P.~Kr{\"{a}}henb{\"{u}}hl, V.~Koltun, Geodesic object proposals, in: Computer
  Vision - {ECCV} 2014 - 13th European Conference, Zurich, Switzerland,
  September 6-12, 2014, Proceedings, Part {V}, 2014, pp. 725--739.

\bibitem{DollarZ13}
P.~Doll{\'{a}}r, C.~L. Zitnick, Structured forests for fast edge detection, in:
  ICCV, 2013, pp. 1841--1848.

\bibitem{ArbelaezPBMM14}
P.~A. Arbel{\'{a}}ez, J.~Pont{-}Tuset, J.~T. Barron, F.~Marqu{\'{e}}s,
  J.~Malik, Multiscale combinatorial grouping, in: CVPR, 2014, pp. 328--335.

\bibitem{SandeUGS11}
K.~E.~A. van~de Sande, J.~R.~R. Uijlings, T.~Gevers, A.~W.~M. Smeulders,
  Segmentation as selective search for object recognition, in: ICCV, 2011, pp.
  1879--1886.

\bibitem{ManenGG13}
S.~Manen, M.~Guillaumin, L.~J.~V. Gool, Prime object proposals with randomized
  prim's algorithm., in: ICCV, 2013.

\bibitem{HeZC04}
X.~He, R.~S. Zemel, M.~. Carreira-Perpiñán, Multiscale conditional random
  fields for image labeling., in: CVPR, 2004, pp. 695--702.

\bibitem{ShottonJC08}
J.~Shotton, M.~Johnson, R.~Cipolla, Semantic texton forests for image
  categorization and segmentation., in: CVPR, 2008.

\bibitem{Farabet2013}
C.~Farabet, C.~Couprie, L.~Najman, Y.~LeCun, Learning hierarchical features for
  scene labeling, IEEE Trans. Pattern Anal. Mach. Intell. 35~(8) (2013)
  1915--1929.
\newblock \href {http://dx.doi.org/10.1109/TPAMI.2012.231}
  {\path{doi:10.1109/TPAMI.2012.231}}.

\bibitem{HoiemEH07}
D.~Hoiem, A.~A. Efros, M.~Hebert, Recovering surface layout from an image,
  International Journal of Computer Vision 75~(1) (2007) 151--172.

\bibitem{Tighe2013}
J.~Tighe, S.~Lazebnik, Superparsing - scalable nonparametric image parsing with
  superpixels., International Journal of Computer Vision 101~(2) (2013)
  329--349.

\bibitem{LadickyRKT09}
L.~Ladicky, C.~Russell, P.~Kohli, P.~H.~S. Torr, Associative hierarchical crfs
  for object class image segmentation, in: {IEEE} 12th International Conference
  on Computer Vision, {ICCV} 2009, Kyoto, Japan, September 27 - October 4,
  2009, 2009, pp. 739--746.

\bibitem{RabinovichVGWB07}
A.~Rabinovich, A.~Vedaldi, C.~Galleguillos, E.~Wiewiora, S.~Belongie, Objects
  in context., in: ICCV, 2007, pp. 1--8.

\bibitem{GalleguillosRB08}
C.~Galleguillos, A.~Rabinovich, S.~Belongie, Object categorization using
  co-occurrence, location and appearance., in: CVPR, 2008.

\bibitem{GalleguillosMBL10}
C.~Galleguillos, B.~McFee, S.~J. Belongie, G.~R.~G. Lanckriet, Multi-class
  object localization by combining local contextual interactions., in: CVPR,
  IEEE, 2010, pp. 113--120.

\bibitem{FelzenszwalbMR08}
P.~F. Felzenszwalb, D.~A. McAllester, D.~Ramanan, A discriminatively trained,
  multiscale, deformable part model., in: CVPR, 2008.

\bibitem{LadickySART10}
L.~Ladicky, P.~Sturgess, K.~Alahari, C.~Russell, P.~H.~S. Torr, What, where and
  how many? combining object detectors and crfs, in: ECCV, 2010.

\bibitem{FlorosRL11}
G.~Floros, K.~Rematas, B.~Leibe, Multi-class image labeling with top-down
  segmentation and generalized robust {\textdollar}p{\^{}}n{\textdollar}
  potentials, in: British Machine Vision Conference, {BMVC} 2011, Dundee, UK,
  August 29 - September 2, 2011. Proceedings, 2011.

\bibitem{LeibeLS08}
B.~Leibe, A.~Leonardis, B.~Schiele, Robust object detection with interleaved
  categorization and segmentation., International Journal of Computer Vision
  77~(1-3) (2008) 259--289.

\bibitem{ArbelaezHGGBM12}
P.~Arbelaez, B.~Hariharan, C.~Gu, S.~Gupta, L.~D. Bourdev, J.~Malik, Semantic
  segmentation using regions and parts, in: CVPR, 2012, pp. 3378--3385.

\bibitem{GuoH12}
R.~Guo, D.~Hoiem, Beyond the line of sight: Labeling the underlying surfaces.,
  in: A.~W. Fitzgibbon, S.~Lazebnik, P.~Perona, Y.~Sato, C.~Schmid (Eds.),
  ECCV, Lecture Notes in Computer Science, Springer, 2012, pp. 761--774.

\bibitem{Tu08}
Z.~Tu, Auto-context and its application to high-level vision tasks, in: CVPR,
  2008.

\bibitem{KappesAHSNBKKLKR13}
J.~H. Kappes, B.~Andres, F.~A. Hamprecht, C.~Schn{\"{o}}rr, S.~Nowozin,
  D.~Batra, S.~Kim, B.~X. Kausler, J.~Lellmann, N.~Komodakis, C.~Rother, A
  comparative study of modern inference techniques for discrete energy
  minimization problems, in: CVPR, 2013, pp. 1328--1335.

\bibitem{VineetWT14}
V.~Vineet, J.~Warrell, P.~H.~S. Torr, Filter-based mean-field inference for
  random fields with higher-order terms and product label-spaces, International
  Journal of Computer Vision 110~(3) (2014) 290--307.

\bibitem{LiuYT11}
C.~Liu, J.~Yuen, A.~Torralba, Sift flow: Dense correspondence across scenes and
  its applications., IEEE Trans. Pattern Anal. Mach. Intell. 33~(5) (2011)
  978--994.

\bibitem{RotherMBK06}
C.~Rother, T.~P. Minka, A.~Blake, V.~Kolmogorov, Cosegmentation of image pairs
  by histogram matching - incorporating a global constraint into mrfs, in:
  CVPR, 2006, pp. 993--1000.

\bibitem{HochbaumS09}
D.~S. Hochbaum, V.~Singh, An efficient algorithm for co-segmentation, in: ICCV,
  2009, pp. 269--276.

\bibitem{BatraKPLC10}
D.~Batra, A.~Kowdle, D.~Parikh, J.~Luo, T.~Chen, icoseg: Interactive
  co-segmentation with intelligent scribble guidance, in: CVPR, 2010.

\bibitem{CollinsXGS12}
M.~D. Collins, J.~Xu, L.~Grady, V.~Singh, Random walks based multi-image
  segmentation: Quasiconvexity results and gpu-based solutions, in: CVPR, 2012.

\bibitem{MengLL12}
F.~Meng, H.~Li, G.~Liu, Image co-segmentation via active contours, in: 2012
  {IEEE} International Symposium on Circuits and Systems, {ISCAS} 2012, Seoul,
  Korea (South), May 20-23, 2012, 2012, pp. 2773--2776.

\bibitem{KimLH12}
E.~Kim, H.~Li, X.~Huang, A hierarchical image clustering cosegmentation
  framework, in: CVPR, 2012, pp. 686--693.

\bibitem{VicenteRK11}
S.~Vicente, C.~Rother, V.~Kolmogorov, Object cosegmentation, in: CVPR, 2011,
  pp. 2217--2224.

\bibitem{MengLLN12}
F.~Meng, H.~Li, G.~Liu, K.~N. Ngan, Object co-segmentation based on shortest
  path algorithm and saliency model, IEEE Transactions on Multimedia 14~(5).

\bibitem{SunP13}
J.~Sun, J.~Ponce, Learning discriminative part detectors for image
  classification and cosegmentation, in: ICCV, 2013, pp. 3400--3407.

\bibitem{DaiWZZ13}
J.~Dai, Y.~N. Wu, J.~Zhou, S.~Zhu, Cosegmentation and cosketch by unsupervised
  learning, in: ICCV, 2013, pp. 1305--1312.

\bibitem{FaktorI13}
A.~Faktor, M.~Irani, Co-segmentation by composition, in: ICCV, 2013, pp.
  1297--1304.

\bibitem{WangRL13}
Z.~Wang, R.~Liu, Semi-supervised learning for large scale image cosegmentation,
  in: Computer Vision (ICCV), 2013 IEEE International Conference on, 2013.

\bibitem{ZhuCZWM13}
H.~Zhu, J.~Cai, J.~Zheng, J.~Wu, N.~Magnenat{-}Thalmann, Salient object cutout
  using google images, in: ISCAS, 2013, pp. 905--908.

\bibitem{RubinsteinJKL13}
M.~Rubinstein, A.~Joulin, J.~Kopf, C.~Liu, Unsupervised joint object discovery
  and segmentation in internet images, in: CVPR, 2013, pp. 1939--1946.

\bibitem{ChaiLZ11}
Y.~Chai, V.~S. Lempitsky, A.~Zisserman, Bicos: {A} bi-level co-segmentation
  method for image classification, in: ICCV, 2011, pp. 2579--2586.

\bibitem{ChaiRLGZ12}
Y.~Chai, E.~Rahtu, V.~S. Lempitsky, L.~J.~V. Gool, A.~Zisserman, Tricos: {A}
  tri-level class-discriminative co-segmentation method for image
  classification, in: ECCV, 2012, pp. 794--807.

\bibitem{MaL13}
T.~Ma, L.~J. Latecki, Graph transduction learning with connectivity constraints
  with application to multiple foreground cosegmentation, in: CVPR, 2013.

\bibitem{Zhu14}
H.~Zhu, J.~Lu, J.~Cai, J.~Zheng, N.~Thalmann, Multiple foreground recognition
  and cosegmentation: An object-oriented crf model with robust higher-order
  potentials, in: WACV, 2014.

\bibitem{ZhuICIP14}
H.~Zhu, J.~Lu, J.~Cai, J.~Zheng, N.~Magnenat{-}Thalmann, Poselet-based multiple
  human identification and cosegmentation, in: ICIP, 2014.

\bibitem{MartinFTM01}
D.~R. Martin, C.~Fowlkes, D.~Tal, J.~Malik, A database of human segmented
  natural images and its application to evaluating segmentation algorithms and
  measuring ecological statistics, in: ICCV, 2001, pp. 416--425.

\bibitem{RussellTMF08}
B.~C. Russell, A.~Torralba, K.~P. Murphy, W.~T. Freeman, Labelme: A database
  and web-based tool for image annotation., International Journal of Computer
  Vision 77~(1-3) (2008) 157--173.

\bibitem{LiCAZ13}
H.~Li, J.~Cai, N.~T.~N. Anh, J.~Zheng, A benchmark for semantic image
  segmentation, in: ICME, 2013, pp. 1--6.

\bibitem{TsaiBMVC10}
D.~Tsai, M.~Flagg, J.~M.Rehg, Motion coherent tracking with multi-label mrf
  optimization, BMVC.

\bibitem{RubioSL12}
J.~C. Rubio, J.~Serrat, A.~M. L{\'{o}}pez, Video co-segmentation, in: ACCV,
  2012, pp. 13--24.

\bibitem{ChiuF13}
W.~chen Chiu, M.~Fritz, Multi-class video co-segmentation with a generative
  multi-video model., in: CVPR, IEEE, 2013, pp. 321--328.

\bibitem{BrostowSFC2008}
G.~J. Brostow, J.~Shotton, J.~Fauqueur, R.~Cipolla, Segmentation and
  recognition using structure from motion point clouds, in: ECCV (1), 2008, pp.
  44--57.

\bibitem{EveringhamGWWZ10}
M.~Everingham, L.~J.~V. Gool, C.~K.~I. Williams, J.~M. Winn, A.~Zisserman, The
  pascal visual object classes {(VOC)} challenge, International Journal of
  Computer Vision 88~(2) (2010) 303--338.

\bibitem{XiaoHEOT10}
J.~Xiao, J.~Hays, K.~A. Ehinger, A.~Oliva, A.~Torralba, {SUN} database:
  Large-scale scene recognition from abbey to zoo, in: CVPR, 2010, pp.
  3485--3492.

\bibitem{Silberman12}
P.~K. Nathan~Silberman, Derek~Hoiem, R.~Fergus, Indoor segmentation and support
  inference from rgbd images, in: ECCV, 2012.

\bibitem{LinMBHPRDZ14}
T.~Lin, M.~Maire, S.~Belongie, J.~Hays, P.~Perona, D.~Ramanan, P.~Doll{\'{a}}r,
  C.~L. Zitnick, Microsoft {COCO:} common objects in context, in: ECCV, 2014,
  pp. 740--755.

\bibitem{Martin:CSD-03-1268}
D.~R. Martin, An empirical approach to grouping and segmentation, Ph.D. thesis,
  EECS Department, University of California, Berkeley (Aug 2003).

\bibitem{Meila03}
M.~Meila, Comparing clusterings by the variation of information, in: COLT,
  2003, pp. 173--187.

\bibitem{UnnikrishnanPH07}
R.~Unnikrishnan, C.~Pantofaru, M.~Hebert, Toward objective evaluation of image
  segmentation algorithms, IEEE Trans. Pattern Anal. Mach. Intell. 29~(6)
  (2007) 929--944.

\bibitem{SturgessALT09}
P.~Sturgess, K.~Alahari, L.~Ladicky, P.~H.~S. Torr, Combining appearance and
  structure from motion features for road scene understanding., in: BMVC,
  British Machine Vision Association, 2009.

\bibitem{ZhangWY10}
C.~Zhang, L.~Wang, R.~Yang, Semantic segmentation of urban scenes using dense
  depth maps., in: K.~Daniilidis, P.~Maragos, N.~Paragios (Eds.), ECCV, Lecture
  Notes in Computer Science, Springer, pp. 708--721.

\bibitem{SilbermanF11}
N.~Silberman, R.~Fergus, Indoor scene segmentation using a structured light
  sensor., in: ICCV Workshops, 2011, pp. 601--608.

\bibitem{SilbermanHKF12}
N.~Silberman, D.~Hoiem, P.~Kohli, R.~Fergus, Indoor segmentation and support
  inference from rgbd images., in: ECCV, 2012, pp. 746--760.

\bibitem{KuttelGF12}
D.~K{\"u}ttel, M.~Guillaumin, V.~Ferrari, Segmentation propagation in imagenet,
  in: ECCV, 2012.

\bibitem{Fanman2014}
F.~Meng, J.~Cai, H.~Li, On multiple image group cosegmentation, in: ACCV, 2014.

\bibitem{HeG14}
X.~He, S.~Gould, An exemplar-based {CRF} for multi-instance object
  segmentation, in: CVPR, 2014, pp. 296--303.

\bibitem{TigheNL14}
J.~Tighe, M.~Niethammer, S.~Lazebnik, Scene parsing with object instances and
  occlusion ordering, in: CVPR, 2014, pp. 3748--3755.

\bibitem{YaoFU12}
J.~Yao, S.~Fidler, R.~Urtasun, Describing the scene as a whole: Joint object
  detection, scene classification and semantic segmentation, in: CVPR, 2012,
  pp. 702--709.

\bibitem{DivvalaHHEH09}
S.~K. Divvala, D.~Hoiem, J.~Hays, A.~A. Efros, M.~Hebert, An empirical study of
  context in object detection, in: CVPR, 2009, pp. 1271--1278.

\bibitem{GuptaSEH11}
A.~Gupta, S.~Satkin, A.~A. Efros, M.~Hebert, From 3d scene geometry to human
  workspace., in: CVPR, 2011, pp. 1961--1968.

\bibitem{VineetWT12}
V.~Vineet, J.~Warrell, P.~H.~S. Torr, Filter-based mean-field inference for
  random fields with higher-order terms and product label-spaces, in: ECCV,
  2012, pp. 31--44.

\bibitem{FarhadiEH10}
A.~Farhadi, I.~Endres, D.~Hoiem, Attribute-centric recognition for
  cross-category generalization., in: CVPR, 2010.

\bibitem{KumarBBN11}
N.~Kumar, A.~C. Berg, P.~N. Belhumeur, S.~K. Nayar, Describable visual
  attributes for face verification and image search., IEEE Trans. Pattern Anal.
  Mach. Intell. 33~(10) (2011) 1962--1977.

\bibitem{LampertNH09}
C.~H. Lampert, H.~Nickisch, S.~Harmeling, Learning to detect unseen object
  classes by between-class attribute transfer., in: CVPR, 2009, pp. 951--958.

\bibitem{TigheL11}
J.~Tighe, S.~Lazebnik, Understanding scenes on many levels., in: D.~N. Metaxas,
  L.~Quan, A.~Sanfeliu, L.~J.~V. Gool (Eds.), ICCV, 2011, pp. 335--342.

\bibitem{ZhengCWSVRT14}
S.~Zheng, M.~Cheng, J.~Warrell, P.~Sturgess, V.~Vineet, C.~Rother, P.~H.~S.
  Torr, Dense semantic image segmentation with objects and attributes, in:
  CVPR, 2014, pp. 3214--3221.

\bibitem{GuptaD08}
A.~Gupta, L.~S. Davis, Beyond nouns: Exploiting prepositions and comparative
  adjectives for learning visual classifiers., in: D.~A. Forsyth, P.~H.~S.
  Torr, A.~Zisserman (Eds.), ECCV, Lecture Notes in Computer Science, Springer,
  2008, pp. 16--29.

\bibitem{KulkarniPODLCBB13}
G.~Kulkarni, V.~Premraj, V.~Ordonez, S.~Dhar, S.~Li, Y.~Choi, A.~C. Berg, T.~L.
  Berg, Babytalk: Understanding and generating simple image descriptions., IEEE
  Trans. Pattern Anal. Mach. Intell. 35~(12) (2013) 2891--2903.

\bibitem{KrizhevskySH12}
A.~Krizhevsky, I.~Sutskever, G.~E. Hinton, Imagenet classification with deep
  convolutional neural networks, in: NIPS, 2012.

\bibitem{ErhanSTA13}
D.~Erhan, C.~Szegedy, A.~Toshev, D.~Anguelov, Scalable object detection using
  deep neural networks, CoRR abs/1312.2249.

\bibitem{ToshevS13}
A.~Toshev, C.~Szegedy, Deeppose: Human pose estimation via deep neural
  networks, CoRR abs/1312.4659.

\bibitem{SzegedyTE13}
C.~Szegedy, A.~Toshev, D.~Erhan, Deep neural networks for object detection, in:
  NIPS, 2013.

\bibitem{GirshickDDM14}
R.~B. Girshick, J.~Donahue, T.~Darrell, J.~Malik, Rich feature hierarchies for
  accurate object detection and semantic segmentation, in: CVPR, 2014, pp.
  580--587.

\bibitem{Freeman2011}
W.~T. Freeman, Where computer vision needs help from computer science, in:
  Proceedings of the Twenty-second Annual ACM-SIAM Symposium on Discrete
  Algorithms, SODA '11, 2011.

\end{thebibliography}

\end{document}